\documentclass{article}

\PassOptionsToPackage{numbers, compress}{natbib}

\usepackage[final]{{neurips_2022}}

\usepackage[utf8]{inputenc} 
\usepackage[T1]{fontenc}    
\usepackage{hyperref}       
\usepackage{url}            
\usepackage{booktabs}       
\usepackage{amsfonts}       
\usepackage{nicefrac}       
\usepackage{microtype}      
\usepackage{xcolor}         

\usepackage{docmute}
\usepackage{{james_package_arxiv}}
\usepackage{mathrsfs}
\usepackage{xcolor}
\usepackage[linesnumbered,ruled,vlined]{algorithm2e}
\usepackage[nottoc]{tocbibind}
\usepackage{natbib}
\usepackage{footnotebackref}
\usepackage{multirow}

\usepackage{thmtools,thm-restate}

\usepackage[totoc]{idxlayout}

\newcommand{\calB}{\mathcal{B}}
\newcommand{\calN}{\mathcal{N}}

\newcommand{\calD}{\mathcal{D}}
\newcommand{\calR}{\mathcal{R}}
\newcommand{\calO}{\mathcal{O}}
\newcommand{\calP}{\mathcal{P}}

\newcommand{\calH}{\mathcal{H}}

\newcommand{\calT}{\mathcal{T}}
\newcommand{\calC}{\mathcal{C}}

\newcommand{\calU}{\mathcal{U}}
\newcommand{\calX}{\mathcal{X}}

\newcommand{\eps}{\epsilon}

\newcommand{\iid}{\overset{iid}{\sim}}

\newcommand{\law}{\mathrm{Distribution}}
\newcommand{\osc}{\mathrm{osc}}

\renewcommand{\P}{\mathds{P}}
\newcommand{\E}{\mathds{E}}

\newcommand{\N}{\mathds{N}}

\newcommand{\scrB}{\mathscr{B}}

\newcommand{\anonymize}[1]{#1}

\newtheorem{lemma}{Lemma}
\newtheorem{theorem}{Theorem}
\newtheorem{proposition}{Proposition}
\newtheorem{corollary}{Corollary}

\theoremstyle{plain}

\newtheorem{assumption}{Assumption}

\newcommand{\nnewline}{}

\SetCommentSty{mycommfont}

\SetKwInput{KwInput}{Input}                
\SetKwInput{KwOutput}{Output}              

\hypersetup{
    colorlinks = true,
    linkbordercolor = {white},
    citecolor = {red},
    linkcolor = {blue}
}

\title{Nonlinear MCMC for Bayesian Machine Learning}
\author{
  James Vuckovic\\
  \texttt{\href{mailto:james@jamesvuckovic.com}{james@jamesvuckovic.com}}
}

\begin{document}
\maketitle

\begin{abstract}
  We explore the application of a nonlinear MCMC technique first introduced in \cite{delmoralnonlin} to problems in Bayesian machine learning. We provide a convergence guarantee in total variation that uses novel results for long-time convergence and large-particle (``propagation of chaos'') convergence. We apply this nonlinear MCMC technique to sampling problems including a Bayesian neural network on CIFAR10.
\end{abstract}

	\section{Introduction}\label{sec:intro}

	Characterizing uncertainty is a fundamental problem in machine learning. It is often desirable for a machine learning model to provide a prediction \emph{and} a measure of how ``certain'' the model is about that prediction. Having access to a robust measure of uncertainty becomes particularly important in real-world, high risk scenarios such as self-driving cars \cite{michelmore2018evaluating,shafaei2018uncertainty,ding2021capture}, medical diagnosis \cite{kompa2021second,alizadehsani2021handling}, and classifying harmful text \cite{he2020towards}.

	However, despite the need for uncertainty in machine learning predictions, it is well known that traditional ML training, i.e. based on optimizing an objective function, frequently does not provide robust uncertainty measures \cite{gal2016dropout}, yielding overconfident predictions for popular neural networks such as ResNets \cite{guo2017calibration}. \footnote{In Appendix~\ref{app:exp_cifar_ood}, we provide an experiment that demonstrates this effect.} An appealing alternative to the traditional optimization paradigm for ML is the Bayesian probabilistic framework, due to its relatively simple formulation and extensive theoretical grounding; see for example \cite{murphy2012machine}.

	From the probabilistic perspective of machine learning \cite{murphy2012machine}, one combines a prior $P(\theta)$ over the parameter space $\theta \in \Theta$ and a likelihood of the data given model parameters $P(\calD|\theta)$ using Bayes' rule to obtain a posterior over the parameters $P(\theta|\calD)\propto P(\calD|\theta)P(\theta)$. The ``traditional'' approach in machine learning is to \emph{optimize} the posterior (or the likelihood) to obtain $\theta^*\in \arg\max P(\theta|\calD)$ and generate predictions via $P(y|x,\calD)=P(y|x,\theta^*)$. However, if we adopt the Bayesian approach, the posterior characterizes the uncertainty about the parameters of the model (i.e. epistemic uncertainty), which can propagate to uncertainty about a prediction by \emph{integration}: $P(y|x,\calD)=\int P(y|x,\theta)P(\theta|\calD)\dd \theta$.  This paper studies the problem of how to approximate this integration with samples from $P(\theta|\calD)$.

	\subsection{Contributions}
	\begin{itemize}
		\item Our main contribution is the novel analysis of a modification of the general nonlinear Markov Chain Monte Carlo (MCMC) sampling method from \cite{delmoralnonlin} to obtain quantitative convergence guarantees in both the number of iterations and the number of samples.
		\item We apply the general results from above to determine the convergence of two specific nonlinear MCMC samplers.
		\item In experiments, we compare these nonlinear MCMC samplers to their linear counterparts, and find that nonlinear MCMC provides additional flexibility in designing sampling algorithms with as good, or better, performance as the linear variety.
	\end{itemize}

	\subsection{Background}
	\paragraph{Bayesian ML \& MCMC.} In Bayesian machine learning, the ``computationally difficult'' step is integration since the integral $\int P(y|x,\theta)P(\dd\theta|\calD)$ is not analytically solvable except for rare cases. In practice, one typically uses a Monte Carlo approximation such as 
	\[
		\int P(y|x,\theta)P(\theta|\calD)\dd \theta \approx \frac{1}{N}\sum_{i=1}^N P(y|x,\theta^i),~~~\text{where}~~~\theta^i\iid P(\theta|\calD)
	\]where the expected error in this approximation is well known to converge to zero like $\calO(1/\sqrt{N})$ by the Central Limit Theorem (CLT). There are various approaches to sampling from $P(\theta|\calD)$, but Markov chain Monte Carlo (MCMC) is perhaps the most widely used. The basic idea of MCMC is to use a Markov transition kernel $\calT$ with stationary measure $P(\theta|\calD)$ to simulate a Markov chain $\theta_{n+1}\sim \calT(\theta_n,\bdot)$ that converges rapidly to $P(\theta|\calD)$. In this case, we can estimate 
	\[
		\int P(y|x,\theta)P(\theta|\calD)\dd \theta\approx \frac{1}{N}\sum_{i=1}^N P(y|x,\theta^i_\infty)\approx \frac{1}{N}\sum_{i=1}^N P(y|x,\theta^i_{n_{\text{sim}}})
	\]where $n_{\text{sim}}$ is some large number of simulation steps and $\{\theta^i_n\}_{n=0}^\infty$ are independent Markov chains governed by $\calT$.

	The basic problem is then to design an efficient transition kernel $\calT$. There is a vast body of literature studying various choices of $\calT$; some well known choices are the Metropolis-Hastings algorithm \citep{metropolis1953equation,hastings1970monte}, the Gibbs sampler \citep{geman1984stochastic}, the Langevin algorithm \citep{doi:10.1063/1.430300,roberts1996exponential,durmus2019high,welling2011bayesian}, Metropolis-Adjusted Langevin \citep{roberts1998optimal,mala}, and Hamiltonian Monte Carlo \citep{duane1987hybrid,betancourt2017conceptual,Neal1992AnIA,neal2012bayesian,girolami2011riemann}. 

	However, there are various challenges in Bayesian ML that make applying these samplers difficult in practice. One challenge is that the posterior $P(\theta|\calD)$ can be highly multimodal \cite{muller1998issues,pourzanjani2017improving}, which makes it difficult to ensure that the Markov chain $\theta_n$ explores all modes of the target distribution. One can combat this issue by employing auxiliary samplers that explore a more ``tractable'' variation of $P(\theta|\calD)$ \citep{10.2307/2670110,habib2018auxiliary}. Other methods that empirically improve posterior sampling quality include tempering \citep{10.1093/gji/ggt342,replica,chandra2019langevin,wenzel2020good}, RMSProp-style preconditioning \citep{li2016preconditioned}, or adaptive MCMC algorithms \citep{andrieu2006ergodicity,andrieu2008tutorial} such as the popular No U-Turn Sampler \citep{hoffman2014no}.

	\paragraph{Nonlinear MCMC.} Another class of powerful MCMC algorithms, which is less-studied in the context of Bayesian ML, arises from allowing the transition kernel $\calT$ to depend on the distribution of the Markov chain as in $\theta_{n+1}\sim \calT_{\law(\theta_n)}(\theta_n,\bdot)$. This approach gives rise to so-called \emph{nonlinear} MCMC since $\{\theta_n\}$ is no longer a true Markov chain. Nonlinear Markov theory is a rich area of research \citep{mckean1966class,meleard1996asymptotic,carrillo2003kinetic,guillin2020uniform,sznitman1991topics,butkovsky2014ergodic} and has strong connections to nonlinear filtering problems \citep{del1997nonlinear,del2001asymptotic}, sequential Monte Carlo \citep{doucet2001introduction,del2006sequential}, and nonlinear Feynman-Kac models \citep{del2004feynman}. One can replace $\law(\theta_n)$, which is often intractable, with an empirical estimate $\law(\theta_n)\approx \frac{1}{N}\sum_{i=1}^N \delta_{\theta^i_n}$ to obtain \emph{interacting} particle MCMC (iMCMC, or iPMCMC) methods; see for example \cite{andrieu2010particle,rainforth2016interacting,delmoralnonlin,clarte2019collective,arbel2019maximum}. \nnewline

	Our view is that nonlinear MCMC offers some appealing features that traditional linear MCMC lacks. One such feature is the ability to leverage \emph{global information} about the state space $\Theta$ contained in $\law(\theta^i_n)$ to improve exploration, a central issue in Bayesian ML. Another feature is the increased flexibility of nonlinear MCMC algorithms, which can be leveraged to correct biases that are introduced by other design decisions in MCMC for Bayesian ML such as tempering. These features will be explored empirically in Section~\ref{sec:exp}.

	However, the theoretical analysis of nonlinear Markov MCMC presents an added difficulty in that the particles $\{\theta^1_n,\dots,\theta^N_n\}$ of an interacting particle system are now \emph{statistically dependent}. This means that, in addition to studying the long-time behaviour which is classical in MCMC \cite{roberts2004general,meyn2012markov}, one must study the large-particle behaviour separately to obtain Monte Carlo estimates since the CLT does not apply. One such large-particle behaviour is the propagation of chaos \cite{sznitman1991topics}, which is the tendency for groups of interacting particles to become independent as the number of particles, $N$, increases; see \cite{sznitman1991topics}. We will need both of these elements --- long-time convergence and propagation of chaos --- to properly characterize the convergence of nonlinear MCMC.

	\paragraph{Other Sampling Methods.}
	Finally, let us mention that MCMC is certainly not the only way to obtain Monte Carlo sample estimates in Bayesian ML; some popular examples include MC dropout \cite{gal2016dropout}, black-box variational inference \cite{ranganath2014black}, and normalizing flows \cite{rezende2015variational,papamakarios2021normalizing}.

	\subsection{Common Notation}
	Let $\calP(\R^d)$ be the set of probability measures on the measurable space $(\R^d,\scrB(\R^d))$. For $\mu\in \calP(\R^d)$ and $f\in \calB_b(\R^d):=\{f:\R^d\to \R\st f\text{ is bounded}\}$, we will denote $\mu(f):=\int f\dd \mu$. If $K:\R^d\times\scrB(\R^d)\to [0,1]$ is a Markov kernel\footnote{i.e. $K(x,\bdot)$ is a probability measure $\forall x\in \R^d$ and $K(\bdot, A)$ is measurable $\forall A\in \scrB(\R^d)$} then we will denote $Kf(x):=\int f(y)K(x,\dd y)$ and $\mu K(\dd y):=\int \mu(\dd x)K(x,\dd y)$. Finally, for $\ovl{y}:=\{y^1,\dots,y^N\}\subset\R^d$, we will denote the empirical measure of $\ovl{y}$ as $m(\ovl{y}):=\frac{1}{N}\sum_{i=1}^N \delta_{y^i}\in \calP(\R^d)$.

	\section{Nonlinear MCMC}\label{sec:Algorithm}
	In this section, we outline the family of MCMC algorithms that will be studied in rest of the work. We will use general notation for simplicity but this difference is merely cosmetic; the ``target distribution'' $\pi$ in this section corresponds directly to $P(\theta|\calD)$ from the previous section.

	\subsection{Nonlinear Jump Interaction Markov Kernels}
	To specify a MCMC algorithm, we must specify the Markov transition kernel. The family of nonlinear Markov kernels that we will be studying was introduced in \cite{delmoralnonlin} and is a mixture of a linear kernel, denoted $K$, and a nonlinear jump-interaction kernel indexed by a probability measure $\eta$, denoted $J_\eta$, to obtain
	\begin{equation}\label{eq:general_nonlin_ker}
		K_\eta(x, \dd y) := (1-\veps) K(x,\dd y) + \veps J_\eta(x,\dd y)
	\end{equation}
	where $\veps\in ]0,1[$ is the mixture hyperparameter. The Markov kernel $K_\eta$ will be the main object of interest throughout this paper. We will give specific examples of $J_\eta$ in Section~\ref{subsec:appl_mcmc}, which were also introduced in \cite{delmoralnonlin}. Despite building on the constructions of \cite{delmoralnonlin}, this work proceeds in some substantially different directions; see Appendix~\ref{app:diffs} for more details.

	\paragraph{Mean Field System.} Now we show how the kernel $K_\eta$ can be used to construct a Markov chain. Following \cite{delmoralnonlin}, we use an auxiliary Markov chain $\{Y_n\}$ with transition kernel $Q$ on the same state space as $K_\eta$ (i.e. $\R^d$) to obtain the nonlinear Markov chain $\{(Y_n,X_n)\}_{n=0}^\infty$ defined by
	\begin{equation}\label{eq:gen_mf_system}
		\rl{
			Y_{n+1} &\sim Q(Y_n,\bdot)\\
			\eta_{n+1} &:= \law(Y_{n+1})\\
			X_{n+1} &\sim K_{\eta_{n+1}}(X_n,\bdot)
		}~~~Y_0\sim \eta_0,~X_0\sim \mu_0
	\end{equation}
	where $\mu_0,\eta_0\in \calP(\R^d)$ are the initial distributions and $\sim$ denotes ``sample from''. One should interpret this as a sequence of steps where first we sample the auxiliary state $Y_{n+1}$ from $Q$, then we obtain the distribution of $Y_{n+1}$ denoted $\eta_{n+1}$, and we use this distribution to index the primary kernel $K_{\eta_{n+1}}$ and obtain a sample $X_{n+1}$. We sample $X_{n+1}$ with probability $(1-\veps)$ from the linear kernel $K$, and with probability $\veps$ it will ``jump'' according to $J_{\eta_{n+1}}(X_n,\bdot)$.  Because the Markov dynamics depend on $\law(Y_{n})$, we call this a ``mean field system''.

	\paragraph{Interacting Particle System.} One issue with the mean field system \eqref{eq:gen_mf_system} is the fact that computing $\law(Y_{n+1})$ is generally impossible except in special cases. Hence, to get a viable simulation algorithm, we must approximate $\law(Y_{n})$, and we do this by replacing $\law(Y_{n})$ with its empirical measure estimated from a set of $N$ particles ${\ovl{Y}_n:=\{Y^1_n,\dots,Y^N_n\}}$ as follows:
	\begin{equation}\label{eq:general_IPS}
		\rl{
			Y^i_{n+1}&\sim Q(Y^i_n,\bdot)\\
			\eta^N_{n+1}&:=m(\ovl{Y}_{n+1})\\
			X^i_{n+1}&\sim K_{\eta^N_{n+1}}(X^i_n,\bdot)
		}~~~Y^i_0\iid \eta_0,~X^i_0\iid \mu_0,~i=1,\dots,N.
	\end{equation}

	\subsection{Application to MCMC}\label{subsec:appl_mcmc}
	Now we detail how to apply $K_\eta$ and the Markov chains \eqref{eq:gen_mf_system} and \eqref{eq:general_IPS} to MCMC. In particular, we must understand how to choose $Q,K,J_\eta$ such that $K_\eta$ will be invariant w.r.t. a target distribution $\pi$. 

	As is usually the case in probabilistic inference problems, we will assume that the target distribution $\pi$ is known only up to a normalizing constant and that it has a density, also denoted $\pi$. We also make the simplifying assumption that $Q$ has an invariant measure $\eta^\star$ (also with density denoted $\eta^\star$) i.e. $\eta^\star Q = \eta^\star$. This is not burdensome; in practice we can, and will, obtain $Q$ from a \emph{linear} MCMC algorithm for some choice of $\eta^\star$. In fact, being able to choose $\eta^\star$ is a powerful design parameter of our methods as we will see in Section~\ref{sec:exp}. We will also assume that the linear kernel $K$ is $\pi$-invariant, i.e. $\pi K=\pi$. 

	To see how we can ensure that $\pi$ is $K_\eta$-invariant, consider the fact that we will design $Q$ s.t. $\eta_n:=\law(Y_n)$ converges to $\eta^\star$. This means we will eventually be sampling from the kernel $K_{\eta^\star}$ and we already have $\pi$-invariance of $K$. Therefore, if we arrange for $J_{\eta^\star}$ to be $\pi$-invariant, $\pi$ will be invariant for $K_{\eta^\star}$ since
	\[
		\pi K_{\eta^\star} = (1-\veps)\pi K + \veps \pi J_{\eta^\star} = (1-\veps)\pi + \veps \pi = \pi.
	\]Intuitively, if the auxiliary chain converges to a steady state and the jumps in that steady state preserve $\pi$ (and $K$ preserves $\pi$), then so will $K_{\eta^\star}$. Now the remaining task is to design nonlinear interaction kernels $J_\eta$ that will yield good performance; we detail two choices below.

	\paragraph{Boltzmann-Gibbs Interaction.} The first choice of $J_\eta$ we will investigate, from \cite{delmoralnonlin}, relies on the Boltzmann-Gibbs transformation \citep{del2004feynman}, which we now explain. Let $G:\R^d \to ]0,\infty[$ be a potential function; then the Boltzmann-Gibbs (BG) transformation is a nonlinear mapping $\Psi_G:\calP(\R^d)\to\calP(\R^d)$ defined by
	\[
		\Psi_G(\mu)(\dd x) := \frac{G(x)}{\mu(G)}\mu(\dd x) ~~~\text{or equivalently}~~~\int f(x)\Psi_G(\mu)(\dd x) := \int f(x)\frac{G(x)}{\mu(G)}\mu(\dd x)
	\]for any $f\in \calB_b(\R^d)$ and whenever $\mu(G)\neq 0$. This transformation has many interesting properties and been extensively studied in \cite{del2004feynman} and related works. 

	To use the BG transformation in MCMC, we will assume that the densities $\pi$ and $\eta^\star$ are positive\footnote{this can be relaxed to $\pi\ll \mu$ and $\mu\ll\pi$} and make the choice that $G(x)$ will be the function
	\begin{equation}\label{eq:pot_fn}
		G(x):=\frac{\pi(x)}{\eta^\star(x)}.
	\end{equation}
	With this choice, we get an interaction kernel $J^{BG}_\eta(x,\dd y):= \Psi_G(\eta)(\dd y)$. We can easily see that
	\[
		\eta^\star(G) = \int \frac{\pi}{\eta^\star}\dd \eta^\star = \int \dd \pi = 1~~~\text{and}~~~\Psi_G(\eta^\star)(\dd x) = \frac{G(x)}{\eta^\star(G)}\eta^\star(\dd x) = \frac{\pi(x)}{\eta^\star(x)}\eta^\star(\dd x) = \pi(\dd x),
	\]i.e. $\Psi_G$ is the multiplicative ``change of measure'' from $\eta^\star$ to $\pi$. Hence the first nonlinear Markov kernel we will investigate is
	\begin{equation}\label{eq:bg_kernel}
		K^{BG}_{\eta}(x,\dd y) := (1-\veps)K(x,\dd y) + \veps \Psi_G(\eta)(\dd y).
	\end{equation}
	From the remarks above, is clear that $\pi$ is $K_{\eta^\star}$-invariant.

	\paragraph{Accept-Reject Interaction.}
	The second choice of jump interaction we will study, also introduced in \cite{delmoralnonlin}, is a type of accept-reject interaction related to the Metropolis-Hastings algorithm. For the \emph{same} choice of potential function $G$ in \eqref{eq:pot_fn}, we can define the acceptance ratio 
	\[
		\alpha(x,y):= 1\wedge \frac{G(y)}{G(x)} = 1\wedge \frac{\pi(y)\eta^\star(x)}{\eta^\star(y)\pi(x)}~~~\text{and the quantity}~~~A_\eta(x):=\int \alpha(x,y)\eta(\dd y)
	\]
	for $\eta\in \calP(\R^d)$. Hence we can define the accept-reject interaction kernel as\footnote{Given $f\in \calB_b(\R^d)$, we can also write this as $J^{AR}_\eta f(x) = \int [f(y) - f(x)]\alpha(x,y)\eta(\dd y) + f(x)$}
	\[
		J^{AR}_\eta(x,\dd y) := \alpha(x,y)\eta(\dd y) + (1-A_\eta(x))\delta_x(\dd y).
	\]
	We can interpret this jump interaction as: starting in state $x$, we jump to a new state distributed according to $\eta(\dd y)$ with probability $\alpha(x,y)$ (i.e. accept the proposed jump) and remain the in current state with probability $1-A_\eta(x)$ (i.e. reject the proposed jump). This is a form of ``adaptive Metropolis-Hastings'' in which the proposal distribution evolves over time as the distribution of the auxiliary Markov chain. Hence we obtain the accept-reject nonlinear jump interaction kernel
	\begin{equation}\label{eq:ar_kernel}
		K^{AR}_\eta(x,\dd y) := (1-\veps)K(x,\dd y) + \veps[\alpha(x,y)\eta(\dd y) + (1-A_\eta(x))\delta_x(\dd y)].
	\end{equation}
	We note that $\pi$ is also $J^{AR}_{\eta^\star}$-invariant; see Proposition~\ref{prop:ar_invar} in Appendix~\ref{app:specific} for a simple calculation.

	\paragraph{Simulation.} Let us note briefly that using both $K^{BG}_\eta$ and $K^{AR}_\eta$ in \eqref{eq:general_IPS} produce interacting particle systems that can, and will, be simulated. The simulation is relatively straightforward, see Appendix~\ref{app:code} for pseudocode implementing the nonlinear MCMC algorithms we have now constructed.

	\section{Convergence Analysis}\label{sec:conv}
	We will now study whether the nonlinear MCMC algorithms based on $K_\eta$ from Section~\ref{sec:Algorithm} --- i.e., the interacting particle system \eqref{eq:general_IPS} with the restrictions on $K,Q,J_\eta$ from Section~\ref{subsec:appl_mcmc} --- will actually converge to the target distribution $\pi$. In other words, we would like to estimate $\|\mu^N_n - \pi\|$ for some suitable notion of distance on $\calP(\R^d)$, where $\mu^N_n:=\law(X^1_n)$ is the distribution of a single particle (it doesn't matter which particle as the $X^i_n$ are \emph{exchangeable}). 

	The nonlinear nature of $K_\eta$ makes this analysis more difficult than of a linear MCMC method. We break the problem into two parts: one studying the convergence of the mean-field system \eqref{eq:gen_mf_system} as the number of steps $n\to \infty$, and one studying the convergence of the interacting particle system \eqref{eq:general_IPS} to the mean field system as the number of particles $N\to \infty$. This will allow us to apply the triangle inequality as follows:
	\[
		\|\mu^N_n - \pi\| \leq \underbrace{\|\mu^N_n - \mu_n\|}_{\text{large-particle convergence}} + \underbrace{\|\mu_n - \pi \|}_{\text{long-time convergence}}
	\]where $\mu_n:=\law(X_n)$ is the distribution of the mean-field system. Crucially, our analysis of the large-particle limit is \emph{uniform} in the number of steps $n$, which will allow us to establish bounds above that hold as $n\to \infty$. The actual result is contained in Theorem~\ref{thm:main_result}.

	While our analysis does not rely on heavy mathematical machinery, to state the full set of conditions and results for long-time and large-particle convergence --- each of which is a substantial result in its own right --- would occupy too much space in the main text. Instead, we will state the main result in Theorem~\ref{thm:main_result}, which is essentially a corollary of the long-time and large-particle analyses Theorems~\ref{thm:gen_conv}~and~\ref{thm:poc_general} in Appendices~\ref{app:longtime} and ~\ref{app:poc} respectively, and below we will sketch the general arguments used in those appendices. The proofs of the main results are in Appendix~\ref{app:main_proofs}. Note that our analysis, and the results we obtain, are novel and not found in \cite{delmoralnonlin}; see Appendix~\ref{app:notation} for an elaboration.

	The following result is stated in terms of the \emph{total variation} metric, defined here for $\mu,\nu\in \calP(\R^d)$ as $\|\mu - \nu\|_{tv}:=\sup_{\|f\|_\infty \leq 1}|\mu(f) - \nu(f)|$ where $\|\bdot\|_\infty$ is the sup-norm on $\calB_b(\R^d)$.

	\begin{restatable}{theorem}{mainthm}[Convergence of Nonlinear MCMC]\label{thm:main_result}
		Under suitable conditions on $K_\eta$ and $Q$, there exist fixed constants $C_1, C_2, C_3>0$, a function $\calR:[0,\infty[\to [1,\infty[$, and $\rho>0$ s.t.
		\[
			\|\mu^N_n - \pi\|_{tv} \leq C_1\frac{1}{N}\calR(1/N) + C_2\rho^n + C_3 n\rho^n.
		\]\hfill $\blacklozenge$
	\end{restatable}
	Let us make a couple of remarks:
	\begin{itemize}
		\item This result shows that, to control the approximation error $\|\mu^N_n - \pi\|_{tv}$, it does not necessarily suffice to run the MCMC algorithm for a large number of steps $n$, since if $n\to \infty$ but $N<\infty$ then our bound on $\|\mu^N_n-\pi\|\not\to 0$. However, this approximation cannot lead to arbitrarily bad results: Theorem~\ref{thm:main_result} provides a quantitative upper bound on how much the MCMC algorithm can be biased. This behaviour is supported empirically; in Figure~\ref{fig:2d_exp_10k} of Appendix~\ref{app:dm_comparison} we provide a clear illustration of how changing $N$ significantly affects the bias of our nonlinear MCMC methods while having no effect on the bias of linear MCMC, as expected. 
		\item This result uses total variation, which is a strong metric that represents a worst-case over \emph{all} bounded functions $f$ (up to rescaling by $\|f\|_\infty$). It is entirely possible that, for many choices of practical $f$, the approximation will be better as we will see empirically.
		\item The constant $\rho$ is, roughly speaking, the slower of the rate of convergence for $Q$ and for $K$. Hence if $K,Q$ are chosen to be efficient samplers with fast convergence, this will result in $\rho\ll 1$ and hence $\mu^N_n$ will also converge quickly.
		\item In our specific samplers $K^{BG}_\eta$ and $K^{AR}$, we will see in Appendix~\ref{app:specific} that $\calR$ is a monotonically increasing function that is lower-bounded by $1$. Hence, as $N\to \infty$, $\frac{1}{N}\calR(\frac{1}{N})\to 0$ as expected.
	\end{itemize}

	A corollary of Theorem~\ref{thm:main_result} is that that we regain a Monte Carlo estimate for the interacting particle system. This result is essentially due to \cite{sznitman1991topics}~Theorem~2.2.

	\begin{restatable}{corollary}{maincoro}[Adapted from \cite{sznitman1991topics},~Theorem 2.2]\label{coro:mc}
		Suppose that Theorem~\ref{thm:main_result} applies to $K_\eta$. Let $\ovl{X}_n:=\{X^1_n,\dots,X^N_n\}$ be the interacting particle system from \eqref{eq:general_IPS}. Then for every $n\in \N$ and $f\in \calB_b(\R^d)$ we have
		\[
			\lim_{N\to \infty}\E\left[\left|\frac{1}{N}\sum_{i=1}^Nf(X^i_n) - \mu_n(f)\right|\right]= 0.
		\]\hfill $\blacklozenge$
	\end{restatable}
	This corollary directly relates to the application of Bayesian ML we are interested in, where we would have $f(\theta)=P(y|x,\theta)$.

	\subsection{Long-Time Bounds}
	There are two main ingredients in the general result on long-time convergence: ergodicity of $K$ and $Q$, and Lipschitz regularity of the interaction kernel $\eta\mapsto J_\eta$. These, along with other technical conditions, produce an estimate of the form $\|\mu_n-\pi\|\leq C_2\rho^n + C_3n\rho^n$ where $\|\bdot\|$ is a weighted total variation norm. The full statement is in Theorem~\ref{thm:gen_conv} of Appendix~\ref{app:longtime}.

	\paragraph{Ergodicity of $K$ and $Q$.} A fundamental requirement of our results is that the \emph{linear} building blocks of $K_\eta$ must converge to their respective stationary measures in an appropriate metric. This type of result is now standard in the Markov chain literature, and we use a result from \cite{hairer2011yet} for $K$ and a result from \cite{delmoralnonlin} for $Q$. The former is actually able to ensure that $K$ is a contraction on $\calP(\R^d)$ w.r.t. a suitable weighted total variation; we use this feature repeatedly in our analysis.

	\paragraph{Lipschitz Regularity of $J_\eta$.} We also need that $\eta\mapsto J_\eta$ is Lipschitz-continuous w.r.t. a weighted total variation norm on Markov kernels (the Lipschitz constant does not have to be $<1$). This regularity is used to translate the convergence of $\eta_n\to \eta^\star$ (as guaranteed by the ergodicity of $Q$) into convergence of $J_\eta\to J_{\eta^\star}$ with the Lipschitz estimate $\|J_{\eta_n}-J_{\eta^\star}\|\lesssim \|\eta_n - \eta^\star\|$. In Appendix~\ref{app:specific}, we verify this analytically for $J^{BG}_\eta$ and $J^{AR}_\eta$, see Lemma~\ref{lem:psi_reg} and \cite{delmoralnonlin}~Proposition~5.3.

	\subsection{Large-Particle Bounds}

	To study the large-particle behaviour, we would like to measure how close subset of $q\in \{1,\dots,N\}$ interacting particles $\{X^1_n,\dots,X^q_n\}\subset \{X^1_n,\dots,X^N_n\}=:\ovl{X}_n$ from \eqref{eq:general_IPS} is to being i.i.d. according to the mean-field measure $\mu_n$. This analysis was pioneered in \cite{sznitman1991topics} under the name ``propagation of chaos'' and formalizes the intuition that, as $N\to \infty$, the influence of any individual particle $\to 0$. 

	To state this more precisely, first note that if we had random variables $Z^i\iid \eta\in \calP(\R^d)$ then the joint distribution of $\ovl{Z}:=\{Z^1,\dots,Z^N\}$ would be $\eta^{\otimes N}$. Hence, as $N\to \infty$ for the interacting particles $\ovl{X}_n$ at time $n$, we expect the distribution of $\{X^1_n,\dots,X^q_n\}$, denoted $\mu^{q,N}_n$, to get closer to the distribution of i.i.d. mean field particles from \eqref{eq:gen_mf_system}, denoted $\mu^{\otimes q}_n$. In other words, we expect $\|\mu^{q,N}_n - \mu^{\otimes q}_n\|_{tv}\to 0$ as $N\to \infty$. The full statement of this result is Theorem~\ref{thm:poc_general} of Appendix~\ref{app:poc}.

	The main condition in our propagation of chaos result is another type of regularity for $\eta\mapsto J_\eta$ which basically requires that, if one approximates a distribution $\eta\in \calP(\R^d)$ by its empirical measure $m(\ovl{Y})$ where $\ovl{Y}:=\{Y^1,\dots,Y^N\}$ and $Y^i\iid \eta$, then $J_{m(\ovl{Y})}\to J_\eta$ as $N\to \infty$. More precisely, there should be a function $\calR:[0,\infty[\to [1,\infty[$, which is ideally nondecreasing, s.t. 
	\[
		|\E[J^{\otimes q}_{m(\ovl{Y})}f(x)] - J^{\otimes q}_{\eta}f(x)| \lesssim \frac{q^2}{N}\calR(q^2/N)~~~\forall x\in \R^d ~\text{and}~ f\in \calB_b(\R^d) \text{ with ``oscillations'' } \osc(f)\leq 1.
	\]The expectation is taken over $\eta^{\otimes N}$, and ``oscillations'' are defined precisely in Appendix~\ref{app:notation}. This inequality is essentially a total variation regularity since we can alternately write {$\|\mu~-~\nu\|_{tv}=\sup\{|\mu(f)-\nu(f)|\st f\in \calB_b(\R^d),~\osc(f)\leq 1\}$} \cite{del2004feynman}.

	\subsection{Analysis of Specific Interaction Kernels}
	The main result Theorem~\ref{thm:main_result} is in terms of conditions on a general $K_\eta$ (i.e. general choices of $K,Q,J_\eta$). To apply this result to the samplers in Section~\ref{sec:Algorithm}, we must establish if these conditions hold for $K^{BG}_\eta$ and $K^{AR}_\eta$. Fortunately this can be done analytically; in Appendix~\ref{app:specific}, we present conditions under which the Lipschitz regularity (see Lemma~\ref{lem:psi_reg} and \cite{delmoralnonlin}~Proposition~5.3) and large-particle regularity (see Corollaries~\ref{coro:bg_unif_poc}~and~\ref{coro:ar_unif_poc}) hold. These results, particularly for $K^{BG}_\eta$, are interesting and rely on novel techniques for controlling the various quantities, sometimes improving over previous methods. Due to space constraints, the results for $K^{BG}_\eta$ and $K^{AR}_\eta$ are in Corollaries~\ref{coro:bg_conv}~and~\ref{coro:ar_conv} from Appendix~\ref{app:specific}.

	\section{Experiments}\label{sec:exp}
	In this section, we detail two experiments designed to explore how one might apply the nonlinear MCMC methods developed in the previous sections to Bayesian machine learning.\footnote{The code used in our experiments can be found at \url{https://github.com/jamesvuc/nonlinear-mcmc-paper}. See also Appendix~\ref{app:code} for a discussion of the implementation details.} Let us state explicitly that our aim is \emph{not} to achieve state-of-the-art with these experiments, nor do we claim that this method will necessarily lead to state-of-the-art results on a particular task. Rather, the aims of these experiments are: to demonstrate that nonlinear MCMC can be applied successfully to large-scale problems; to compare linear vs nonlinear methods to understand what benefits and drawbacks nonlinear MCMC offers compared to linear MCMC in practice; and to develop some recipes for choosing the various hyperparameters and samplers that determine a nonlinear MCMC method.

	\subsection{Two-Dimensional Toy Experiments}\label{subsec:2d}
	First, we use a toy setting of two-dimensional distributions to compare the relative benefits of linear and nonlinear MCMC. A benefit of this simple setting is that the multimodal toy distributions can be exactly sampled. This gives us the opportunity to quantify the quality of our samples via an unbiased estimator of the Maximum Mean Discrepancy (MMD) metric on $\calP(\R^d)$ \cite{gretton2012kernel}. This approach stands in contrast to many previous works, which use simplistic distributions (e.g. Gaussians) with analytically tractable statistics to measure quality. See Appendix~\ref{app:exp_2d} for an overview of our methodology.

	\paragraph{Setup.} Our setup will compare the Metropolis Adjusted Langevin Algorithm (MALA) \cite{roberts1998optimal} (see Appendix~\ref{app:mcmc_algs} for an overview of MALA) with the nonlinear BG and AR samplers using MALA for the kernels $K,Q$ in our nonlinear setup from Section~\ref{sec:Algorithm}. This will allow us to examine the effects of the nonlinearity in $K^{AR}_\eta$ and $K^{BG}_\eta$ while controlling for the type of sampler used and its hyperparameters.

	The main difference between the linear and nonlinear algorithms, aside from the interaction itself, is the extra ``design knob'' to control in the form of the choice of the auxiliary density $\eta^\star$. Below, we show how one can use $\eta^\star$ to incorporate additional insight to guide the sampling, such as regions of the state space to explore. In our experiments, we choose $\eta^\star$ to be a centered, 2-dimensional Gaussian with a large variance ($\Sigma=4I_2$ for the circular MoG and two-rings densities, and $\Sigma=20I_2$ for the grid MoG). This conveys ``coarse-grained'' information of roughly where the support of the target density is located -- in this case, a neighbourhood of $(0,0)$. See Table~\ref{tab:2d_settings} in Appendix~\ref{app:exp_2d} for a full account of our experimental settings.

\begin{figure}[ht!]
		\centering
		\begin{tabular}{L{3.5cm} L{3.5cm} L{3.5cm}}
			\includegraphics[scale=0.175]{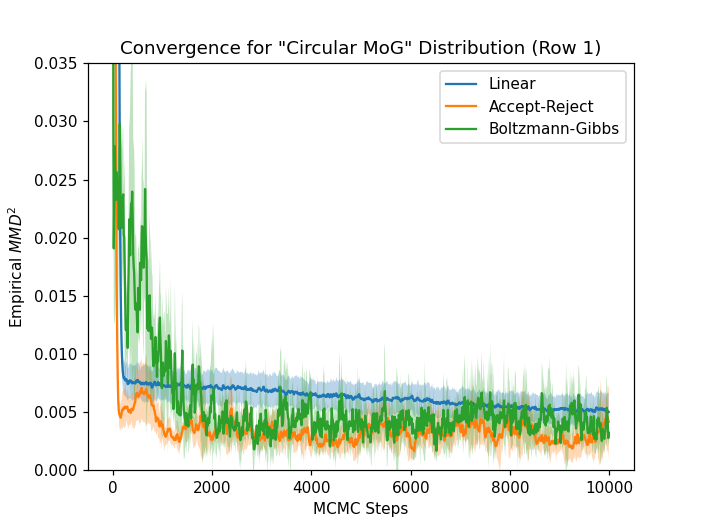} & \includegraphics[scale=0.175]{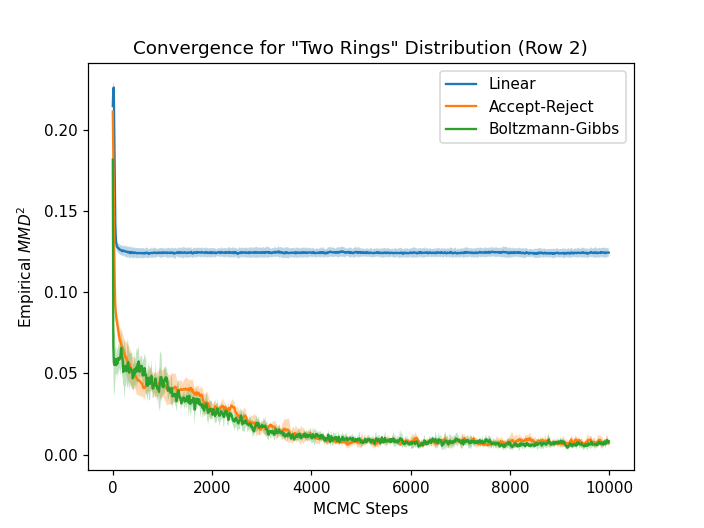} & \includegraphics[scale=0.175]{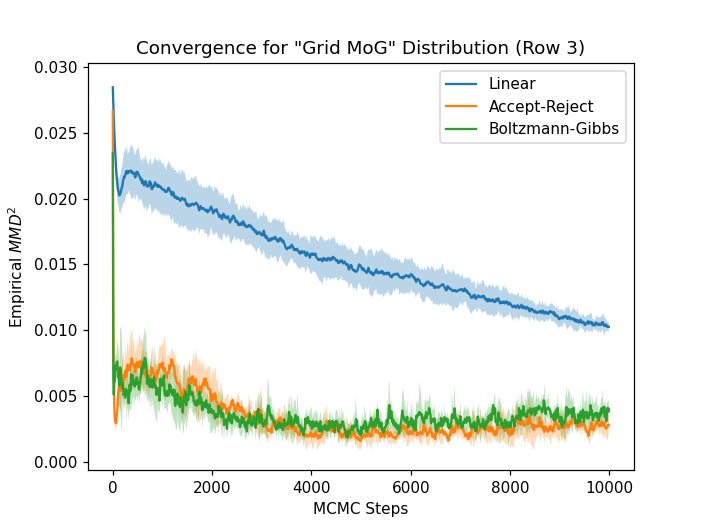}
		\end{tabular}
		\begin{tabular}{C{1.5cm} C{1.5cm} C{1.5cm} C{1.5cm} C{1.5cm}}
			\includegraphics[width=1.75cm, height=1.75cm]{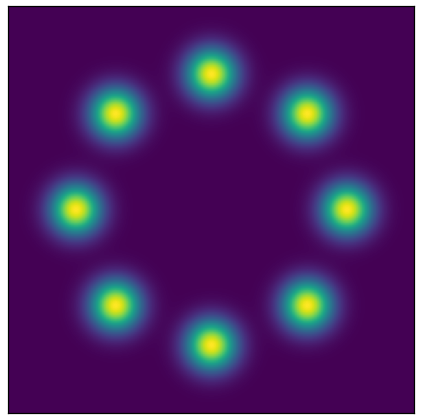} & 
			\includegraphics[width=1.75cm, height=1.75cm]{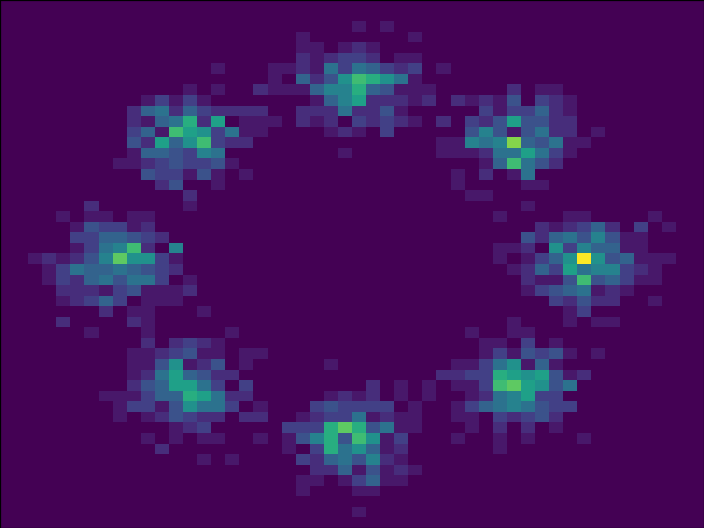} & 
			\includegraphics[width=1.75cm, height=1.75cm]{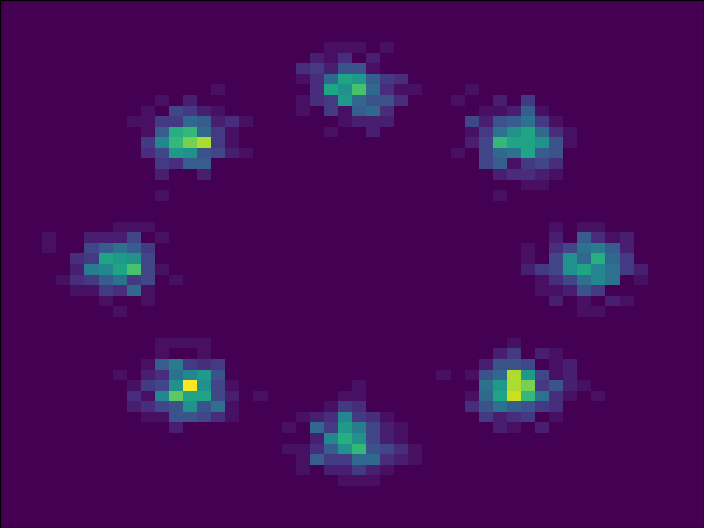} & 
			\includegraphics[width=1.75cm, height=1.75cm]{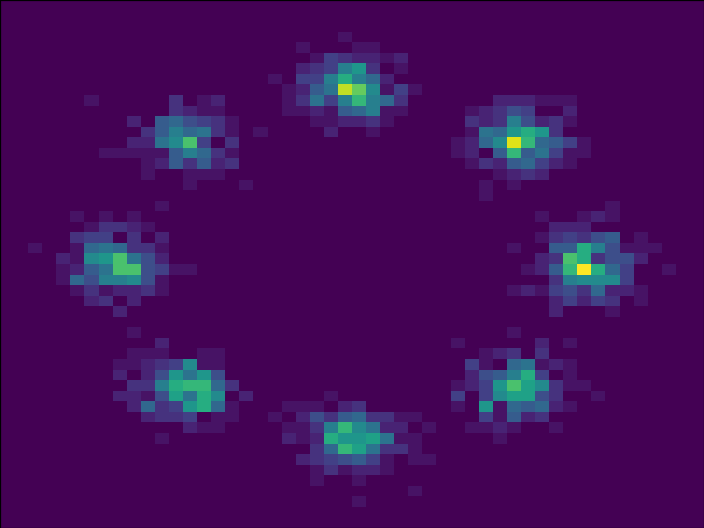} &
			\includegraphics[width=1.75cm, height=1.75cm]{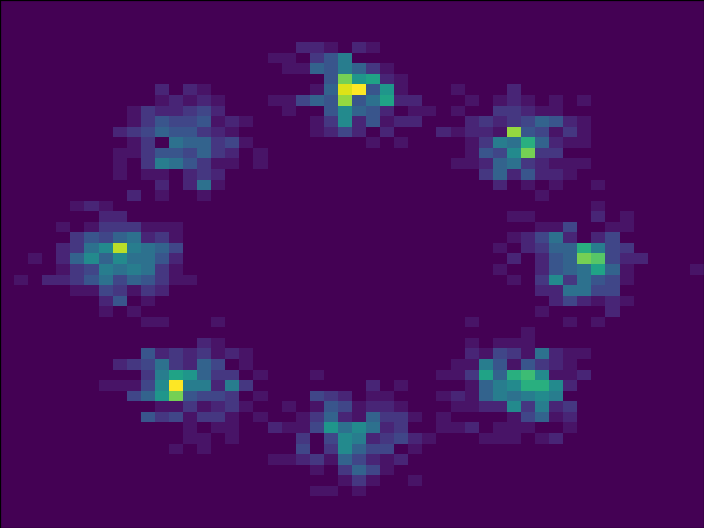} 
			 \\
			\includegraphics[width=1.75cm, height=1.75cm]{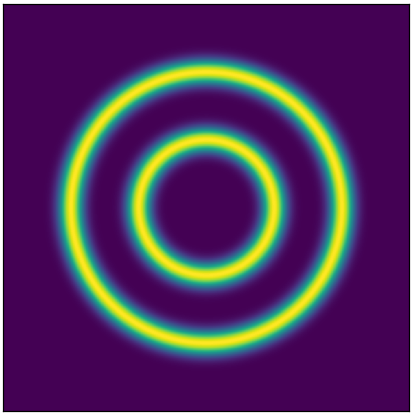} & 
			\includegraphics[width=1.75cm, height=1.75cm]{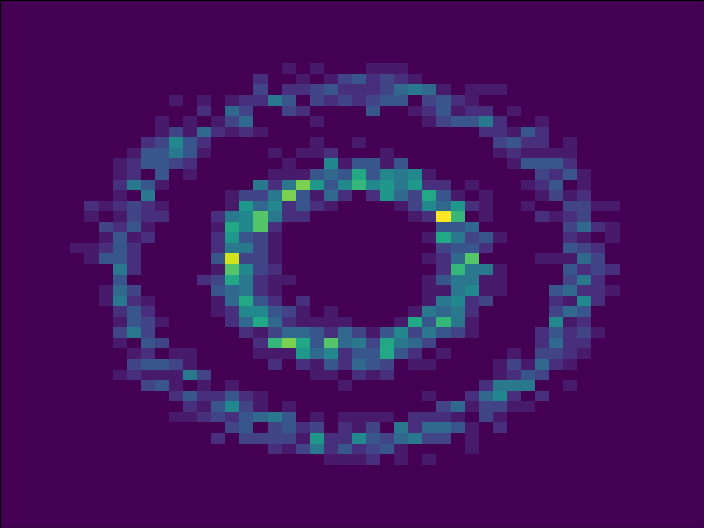} & 
			\includegraphics[width=1.75cm, height=1.75cm]{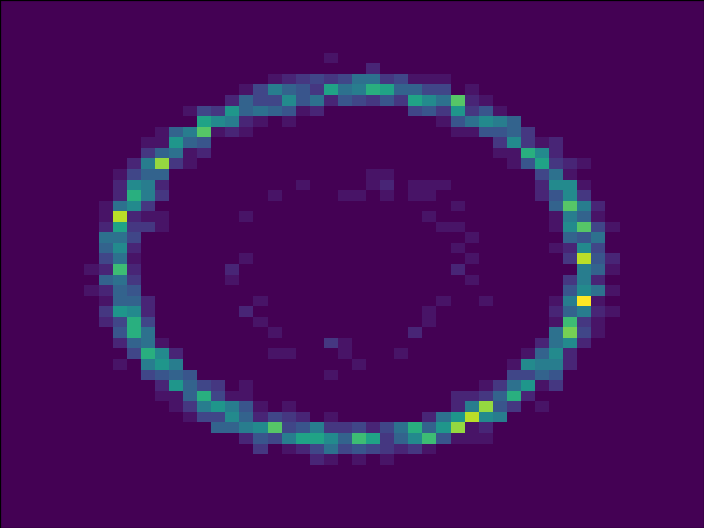} & 
			\includegraphics[width=1.75cm, height=1.75cm]{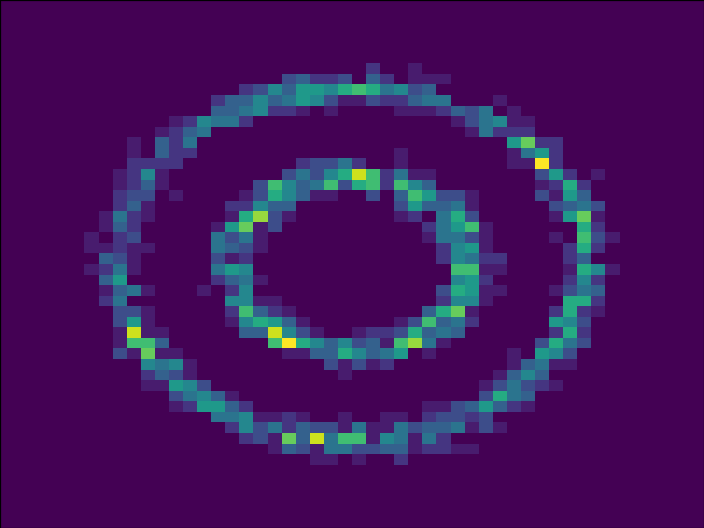} &
			\includegraphics[width=1.75cm, height=1.75cm]{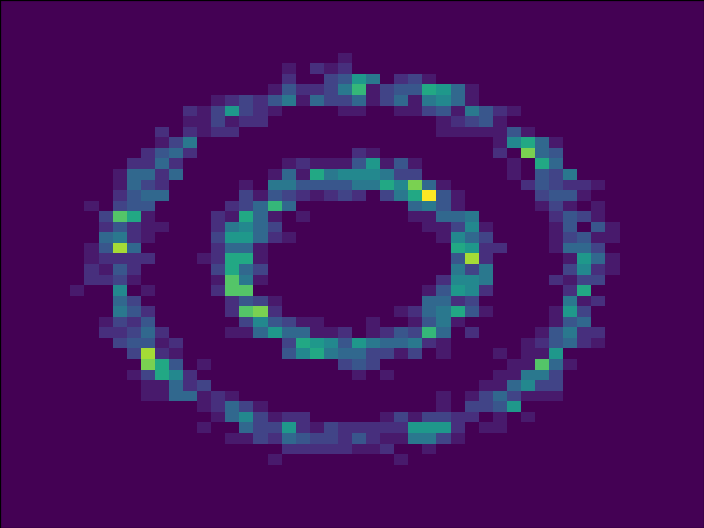}
			\\
			\includegraphics[width=1.75cm, height=1.75cm]{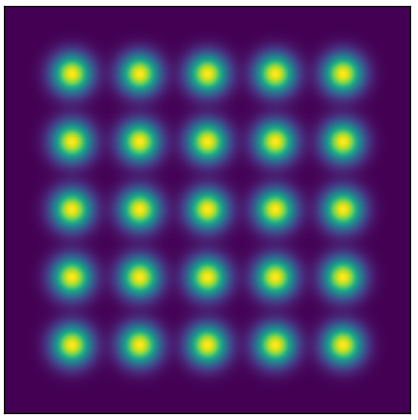} & 
			\includegraphics[width=1.75cm, height=1.75cm]{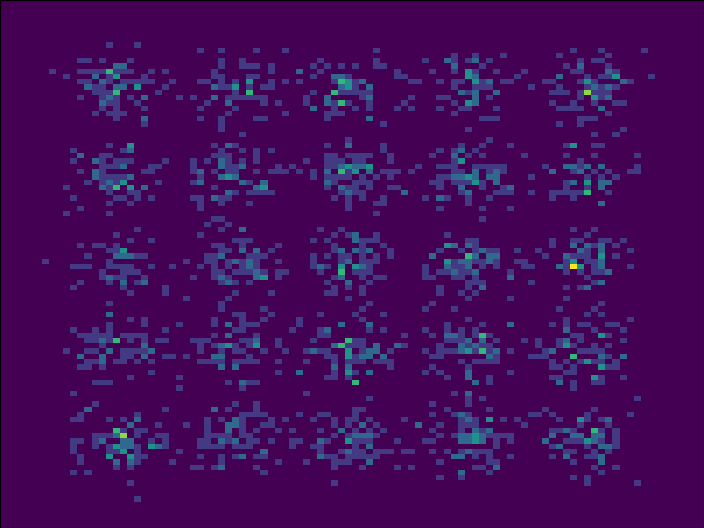} & 
			\includegraphics[width=1.75cm, height=1.75cm]{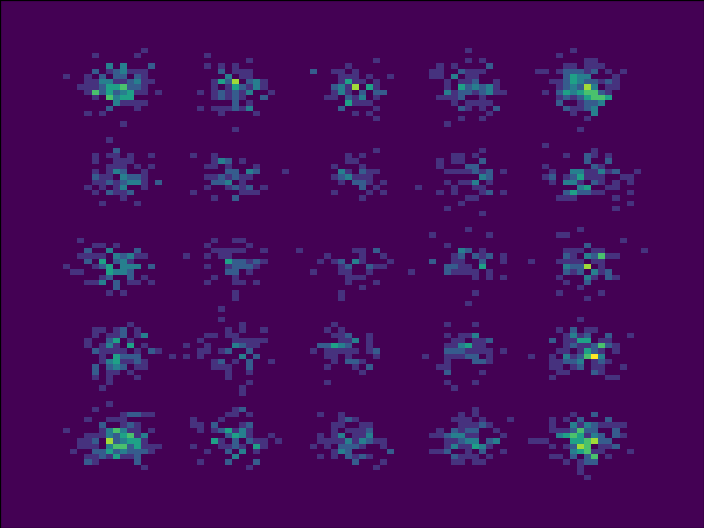} & 
			\includegraphics[width=1.75cm, height=1.75cm]{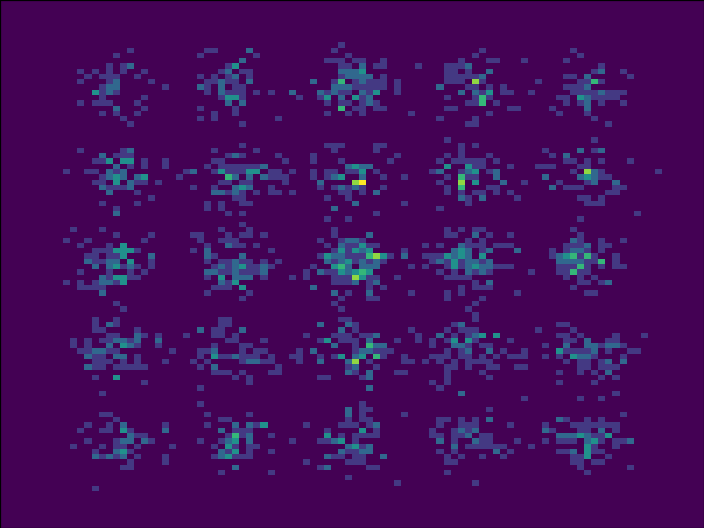} &
			\includegraphics[width=1.75cm, height=1.75cm]{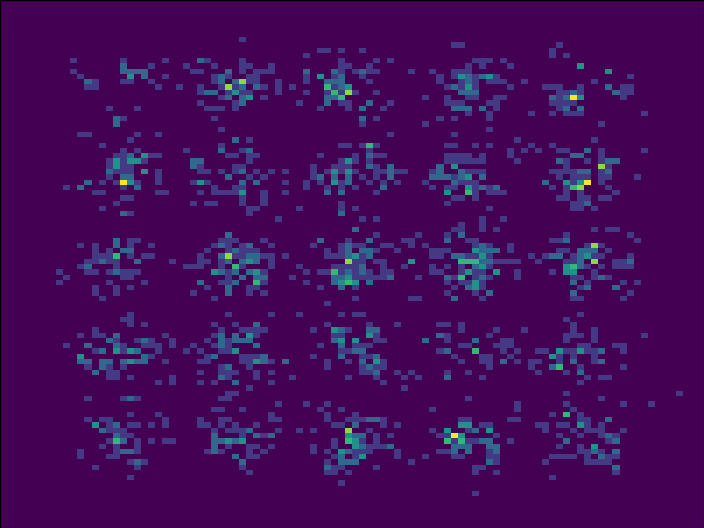} 
			\\
			Target Density & Ground Truth & Linear (MALA) & Nonlinear (AR) & Nonlinear (BG)
		\end{tabular}
		\caption{Visualizations of the 2d experiment. The top row shows the empirical MMD-squared plotted over number of sampled steps, where the shaded region is $\pm 1$ standard deviation with $5$ independent runs. The bottom three rows show histograms for the $N=2000$ samples of the Circular Mixture of Gaussians (MoG) density \cite{stimper2021resampling}, the Two Rings density \cite{stimper2021resampling}, and the Grid Mixture of Gaussians density \cite{zhang2019cyclical} respectively.}
		\label{fig:2d_toy}
	\end{figure}

	\paragraph{Results.} From Figure~\ref{fig:2d_toy}, we see that having a simple auxiliary density with good coverage of the support of the target distribution is quite helpful. In all three examples, one or both of the nonlinear samplers outperformed the equivalent linear sampler in the empirical MMD metric. For the two most challenging densities, the ``two rings'' and ``grid MoG'' distributions, the improved exploration is particularly evident. We also include an analysis of the runtime of the algorithms in Appendix~\ref{app:exp_2d_runtime}.

	\paragraph{Comparison With \cite{delmoralnonlin}.}
	We also compared the performance of our methods with those of \cite{delmoralnonlin} in this two-dimensional toy setting. See Appendix~\ref{app:dm_comparison} for an overview of the results; they support all of the theoretical considerations and design principles we have introduced in this paper.

	\subsection{CIFAR10}

	\subsubsection{Setup}
	To examine the properties of the nonlinear sampler outside of a toy setting, we have also implemented a Bayesian neural network on the CIFAR10 dataset. We use a likelihood $P(y|x,\theta)$ parameterized by a ResNet-18 convolutional neural network \cite{he2016deep} and a Gaussian prior $P(\theta)$ on the parameters of this neural network which are combined to form a posterior $P(\theta|\calD_{\text{train}})\propto P(\theta)\prod_{(x_i, y_i)\in \calD_{\text{train}}} P(y_i|x_i,\theta)$. The goal is to sample $\theta^{i},~i=1,\dots,N$ from this posterior. See Table~\ref{tab:cifar_settings} in Appendix~\ref{app:exp_cifar} for a full account of the experimental settings.

	To deal with the fact that this sampling problem is very high-dimensional $(d\approx 11\mathrm{M})$ and $|\calD_{\text{train}}|$ is large $(|\calD_{\text{train}}|=60,000)$ we use a variety of techniques:
	\begin{enumerate}
		\setlength{\itemsep}{-0.5mm}
		\item We sample minibatches $\wat{\calD}$ of size 256 to obtain the surrogate target density  $P(\theta|\wat{\calD})$ \cite{welling2011bayesian}.
		\item We use an RMSProp-like ``preconditioned'' Langevin algorithm, called RMS-Langevin or RMS-ULA, as in \cite{li2016preconditioned} for the auxiliary sampler $Q$; see Appendix~\ref{app:mcmc_algs} for details on this sampler. As shown in \cite{li2016preconditioned}, this sampler is biased.
		\item We use tempering, wherein we aim to sample from $\pi$ or $\eta\propto P(\theta|\calD_{\text{train}})^{1/\tau}$ where $\tau$ is a small number. This substantially improves mixing for hard-to-sample distributions such as $P(\theta|\calD_{\text{train}})$ at the cost of bias since we are no longer sampling from the true posterior \cite{wenzel2020good}.
	\end{enumerate}

	Using our nonlinear algorithm presents a novel opportunity to correct the bias introduced by tempering. For our experiments, we pick $\eta^\star\propto P(\theta|\calD_{\text{train}})^{1/\tau}$ and $\pi=P(\theta|\calD_{\text{train}})$. This means that the auxiliary chain $Y_n$ explores a tempered version of the target, whereas the target chain $X_n$ (in theory) explores the true target distribution. This is a novel strategy that is made possible by being able to select $\eta^\star$ almost independently of the target $\pi$. We study the case when $\pi$ is tempered as well.

	For our experiments, we use the RMS-Langevin sampler as the baseline, and we also use it for the auxiliary sampler $Q$. For the target sampler, it is not possible to use the RMS-Langevin algorithm because the smoothed square-gradient estimate is incompatible with the discontinuities (i.e. jumps) introduced by the nonlinear interaction. Instead, for the linear sampler $K$ we use the unadjusted Langevin algorithm, ULA, \cite{durmus2019high} (see Appendix~\ref{app:mcmc_algs}). We investigate both test accuracy and calibration error \cite{guo2017calibration} to assess performance.

	\vspace{-3mm}
	\begin{table}[ht!]
		\centering
		\caption{Results for CIFAR10 experiments. $\pm$ represents 1 standard deviation on $5$ random seeds. The tempered results are using $\tau=10^{-4}$. See Appendix~\ref{app:exp_cifar} for an overview of expected calibration error. We also compute the maximum calibration error in Appendix~\ref{app:exp_cifar}. All Expected Calibration Error numbers are multiplied by $10^2$ in this table.}
		\label{tab:cifar_res}
		\vspace{2mm}
		\begin{tabular}{c c c c c c c c c}
			& \multicolumn{2}{c}{\textbf{Test Accuracy} ($\uparrow$)} & \multicolumn{2}{c}{\textbf{Expected Calibration Error} ($\downarrow$)}\\
			Algorithm & Non-Tempered & Tempered & Non-Tempered & Tempered \\
			\hline
			Linear & $85.01_{\pm 0.10}$ & $85.01_{\pm 0.19}$ & $0.24_{\pm 0.02}$ & $0.26_{\pm 0.014}$ \\ 
			Nonlinear (BG) & $84.28_{\pm 0.28}$ & $84.74_{\pm 0.08}$ & $0.14_{\pm 0.03}$ & $0.16_{\pm 0.03}$ \\
			Nonlinear (AR) & \emph{Diverged}  & $84.67_{\pm 0.23}$ & \emph{Diverged} & $0.15_{\pm 0.05}$ \\
			\hline
		\end{tabular}
	\end{table}
	
	\begin{figure}[ht!]
		\centering
		\begin{tabular}{c c}
			\includegraphics[scale=0.2]{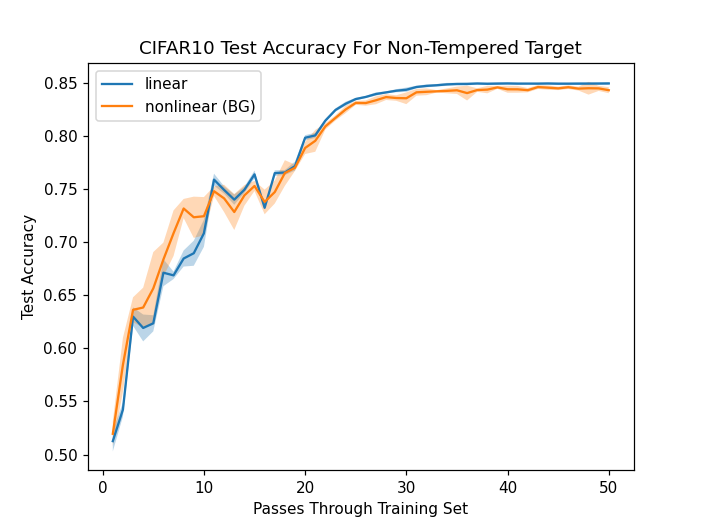} 
			& 
			\includegraphics[scale=0.2]{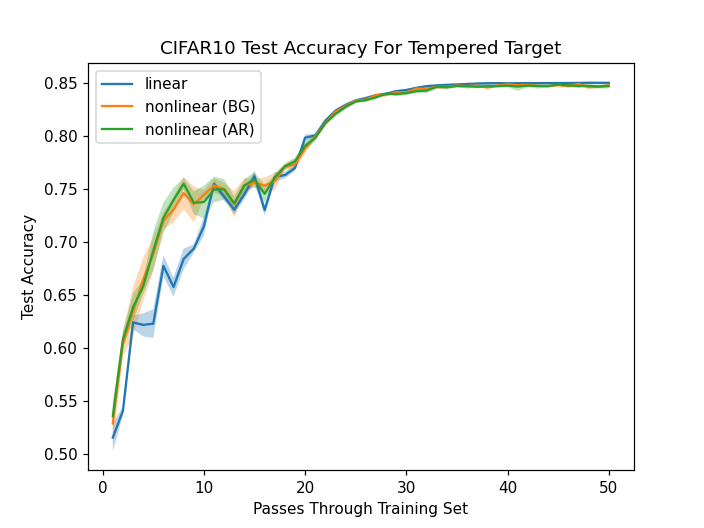}
		\end{tabular}
		\caption{Evaluation of test accuracy during sampling for CIFAR10. The shaded areas represent $\pm 1$ standard deviation for $5$ random seeds. For readability, we omit the AR interaction curve on the non-tempered result since it diverged and it distorts the scale of the plot. For completeness, all the curves for the non-tempered case are plotted in Appendix~\ref{app:cifar10_addtl_plots}, Figure~\ref{fig:diverged}.}
		\label{fig:cifar_explore}
	\end{figure}

	\subsubsection{Results}

	\paragraph{Linear vs Nonlinear.}
	From Table~\ref{tab:cifar_res}, we see that the linear (RMS-Langevin) sampler has slightly higher, but comparable test accuracy to the nonlinear samplers. This is likely because RMS-Langevin algorithm has better stability around the regions of high probability due to its adaptive stepsize scaling. However, from Figure~\ref{fig:cifar_explore}, we see that for both the tempered and non-tempered cases, the nonlinear interaction appears to benefit during early exploration. This is an expected and desired property of these nonlinear samplers, which incorporate global information about the sampler state (in this case, the relative potential $G$ of each auxiliary chain's state) and are able to emphasize those states with higher probability. However, the linear method is able to eventually explore the relevant regions of the state space, and the difference disappears. See Appendix~\ref{app:cifar10_runtime} for a comparison linear vs nonlinear performance scaled by the number of gradient evaluations.

	\paragraph{Tempering Vs Non-Tempering.}

	The linear MCMC sampler is always tempered in our experiments so there should not be any statistically significant difference in the linear case. For the nonlinear sampler, the tempered version has slightly higher accuracy; this trend is also observed in \cite{wenzel2020good}. On the other hand, the calibration errors are the same for both tempered and non-tempered variants. This is somewhat surprising, given the aggressive tempering used, and one would expect that this reduces the variance of the posterior estimate.\footnote{As $\tau\to 0$, this $\theta\sim P(\theta|\calD_{\text{train}})^{1/\tau}$ converges to the maximum \emph{a posteriori} estimate with zero variance.} This observation can perhaps be explained by the fact that we are using $N=10$ samples which may not be enough to accurately change the tempered auxiliary distribution $\eta^\star$ into the non-tempered primary distribution $\pi$ in the jump interaction.

	\paragraph{Calibration.}
	Considering the expected calibration error (ECE) \cite{guo2017calibration}, in Table~\ref{tab:cifar_res} we see that the nonlinear method has statistically significantly lower ECE $(p\ll 0.05)$ compared to the linear method. We hypothesize that this is due to a tension between the RMS scaling of the gradient which improves the efficiency of each MCMC step but at the cost of bias, which may be measurable in the form of calibration error. The Langevin algorithm is also biased, but is generally known to have good convergence properties \cite{durmus2019high} and much less is known about the RMS-Langevin variant. By using our nonlinear setup, we are able to aggressively explore the auxiliary distribution without sacrificing calibration on the target distribution.

	\section{Conclusion}
	\vspace{-2mm}
	In this paper, we have studied the theoretical and empirical properties of nonlinear MCMC methods. We have obtained powerful theoretical results to characterize the convergence of our MCMC methods, and we have applied these methods to Bayesian neural networks. The results on BNNs are comparable to, but not better than, the linear methods we studied. We hypothesize that this is because more investigation into choosing the best auxiliary density $\eta^\star$ is required; our choice is simplistic and may not be optimal. This hypothesis is supported by our toy experiments, which show significant improvement when $\eta^\star$ is able to incorporate some additional insight into the problem. How to do this in high dimensions is an exciting direction for future research.

	\paragraph{Broader Impact.}
	There are benefits and drawbacks to the nonlinear MCMC methods we describe. The benefits are mainly that properly accounting for uncertainty in machine learning will lead to better real-world outcomes for high-value scenarios such as self-driving cars or medical imaging. The drawbacks are that MCMC methods require $\calO(N)$ storage and computations relative to the $\calO(1)$ for deterministic methods; in fact, our nonlinear method would require $2N$ resources compared with $N$ for a linear method (see also Appendices~\ref{app:exp_2d_runtime}~and~\ref{app:cifar10_runtime}). If our algorithms were applied to a large swath of ML ``as is'', this would mean a substantial increase in the energy consumption required for experimentation and deployment, worsening an already substantial issue in the field.

\bibliographystyle{IEEEtraN}
\bibliography{./aux_nonlin_mcmc.bib}

\newpage
\appendix

	\section{Pseudocode \& Numerical Implementation Details}\label{app:code}

	\subsection{Simulation Algorithm}\label{app:diffs}
	The method of simulating $K_\eta$ is where this work and \cite{delmoralnonlin} diverge: we will provide an algorithm that uses a fixed number of samples $N\in \N$ to simulate the IPS \eqref{eq:general_IPS}, whereas \cite{delmoralnonlin} investigates an algorithm in which the empirical measure is formed from all the \emph{past samples}. 

	While the algorithm in \cite{delmoralnonlin} yields asymptotically unbiased estimate of the target measure as $n\to \infty$ (for $K^{AR}_\bdot$ at least, see \cite{delmoralnonlin}), their MCMC algorithm's memory complexity increases linearly with time. This behaviour is not well-suited to large-scale software implementations, which typically favour fixed-size ``batches'' of computations. Additionally, a practitioner cannot add more samples to increase the sampling accuracy in a fixed time frame.

	By contrast, Algorithm~\ref{alg:sampling} below uses a fixed number of particles $N$ throughout the lifetime of the algorithm. Theorem~\ref{thm:main_result} indicates that this produces a biased estimate of the target measure $\pi$, but that this bias can be controlled both in the number of particles, $N$, and the number of steps, $n$. In exchange, our algorithm can be efficiently implemented in vectorized computing frameworks, as we demonstrate in Section~\ref{sec:exp}. In Appendix~\ref{app:dm_comparison}, we investigate the differences in empirical performance between the algorithms in \cite{delmoralnonlin} and Algorithm~\ref{alg:sampling}; those findings unambiguously support this discussion.

	\begin{algorithm}
	  \DontPrintSemicolon
	  \KwInput{Initial samples $X^i_0\iid \mu_0,~Y^i_0\iid \eta_0,~i=1,\dots,N$}
	  \KwInput{Primary and auxiliary Markov kernels $K,Q$ resp. and jump kernel $J_\eta$}
	  \KwInput{Number of iterations, $n_{\text{sim}}$, jump probability $\veps$}
	  \KwOutput{Collection of samples $\{X^1_{n_{\text{sim}}},\dots,X^N_{n_{\text{sim}}}\}$.}
	  
	  \For{$n=0,\dots,n_{\text{sim}}-1$}{ 
	      \tcp{Sample auxiliary Markov chain}
	      \For{$i=1,\dots,N$}{
	        $Y^i_{n+1}\sim Q(Y^i_n, \bdot)$
	      }
	      \tcp{Sample nonlinear Markov chain}
	      \For{$i=1,\dots,N$}{
	        
	        $B^i\sim \text{Bernoulli}(\veps)$ \tcp{Sample binary jump/no jump random variable}
	        \If{$B^i==0$}
	        {
	          Set $X^i_{n+1}\sim K(X^i_n, \bdot)$ \tcp{No jump; evolve according to $K$}
	        }
	        \Else{
	          Set $\ovl{Y}_{n+1}=\{Y^1_{n+1},\dots,Y^N_{n+1}\}$\\
	          Sample $X^i_{n+1}\sim J_{m(\ovl{Y}_{n+1})}(X^i_n,\bdot)$ \tcp{Jump; sample a new position.}
	        }
	      }
	    }
	\caption{Sampling from a nonlinear Markov chain with transition $K_\eta$. }
	\label{alg:sampling}
	\end{algorithm}

	Additionally, let us note that the decision to use an auxiliary Markov chain is highly pragmatic. It is possible to develop ``autonomous'' nonlinear MCMC algorithms of the form $\wtilde{\mu}_{n+1}=K_{\wtilde{\mu}_n}$ (see e.g. \cite{del2004feynman} Ch. 5) with strong theoretical guarantees but bad empirical performance. A primary reason for this is \emph{sample degeneracy}; due to the properties of nonlinear interaction, in an autonomous nonlinear Markov kernel, quite often a single particle $X^i_n$ will be given a large potential $G(X^i_n)$ which results in the jump interaction being concentrated on a single point. From this point onwards, the algorithm will be unable to generate enough diversity within its particles to correctly estimate the variance of the target measure. Due to the close relationship of nonlinear MCMC and nonlinear filtering, these issues have been studied in many settings such as \cite{pitt1999filtering,andrieu2010particle} and using auxiliary dynamics is indeed a common solution.

	\subsection{Efficient Software Implementation}
	In Algorithm~\ref{alg:sampling}, we gave a high-level pseudocode implementation of the nonlinear sampler. In this implementation, for the sake of notational clarity, we used \verb|for| loops to sample the individual particles. However, with modern single instruction multiple data (SIMD) computing frameworks such as GPU accelerators, this is highly inefficient; one can, and should, parallelize all \verb|for| loops except the outer ``time'' loop in Algorithm~\ref{alg:sampling}.

	We have done this using the variety of powerful tools provided by the \verb|JAX| library \citep{jax2018github} within Python such as \verb|vmap| which allows for automatic vectorization, \verb|jit| compilation, and seamless targeting of GPU accelerators. \verb|vmap| is particularly useful for automatic batch-wise and sample-wise vectorization in the case of stochastic gradient MCMC while writing functions in-terms of single inputs and outputs. We have also leveraged the library of linear MCMC algorithms provided by the \verb|jax-bayes| library \anonymize{\url{https://github.com/jamesvuc/jax-bayes}}. All experiments were run on a single Titan RTX 3090 GPU; the code can be found at \anonymize{\url{https://github.com/jamesvuc/nonlinear-mcmc-paper}}.

	\section{Linear MCMC Sampling Algorithms}\label{app:mcmc_algs}
	Markov chain Monte Carlo algorithms form an integral part of Bayesian inference. In this section, we will review the basic MCMC algorithms used in this paper. In each case, we assume a $C^1$, strictly positive target density $\pi$ known only up to a normalizing constant. 

	\subsection{Unadjusted Langevin Algorithm}
	Consider the (overdamped) continuous Langevin diffusion \cite{pavliotis2014stochastic}
	\[
		\dd X_t = \nabla \log\pi(X_t)\dd t + \sqrt{2}\dd B_t
	\]where $B_t$ is a standard Brownian motion. This is a fundamental stochastic process with far reaching consequences in many areas of math; in particular, the Langevin diffusion is a Markov process with $\pi$ as a stationary measure \cite{pavliotis2014stochastic}. We obtain the unadjusted Langevin algorithm, or ULA, by applying an Euler-Maruyama discretization with stepsize $\delta>0$ to the SDE above yielding \cite{pavliotis2014stochastic}
	\[
		X_{n+1} = X_n + \delta \nabla \log \pi(X_n) + \sqrt{2\delta}Z_n,~~~Z_n\sim \calN(0,I_d).
	\]This is a popular and well-studied algorithm for MCMC, although it is known to be \emph{biased} in the sense that the sationary measure of this Markov chain is not $\pi$ \cite{talay1990expansion}. However, this bias converges to 0 as $\delta\to 0$ \cite{durmus2019high}.

	\subsection{Metropolis-Adjusted Langevin Algorithm} 
	The Metropolis-Adjusted Langevin algorithm (MALA) \cite{roberts1998optimal,mala}, like all basic Metropolis-Hastings MCMC methods, consists of two steps: a proposal step and an accept step. In the proposal step, a candidate next state starting from the current state $X_n$ is sampled according to the Langevin dynamics above, i.e. $\wtilde{X}_{n+1} = X_n + \delta\nabla \log \pi(X_n) \sqrt{2\delta}Z_n$. We will write $q(y|x)$ for the proposal distribution; in this case $q(y|x)=\calN(x+\delta \nabla \log\pi(x); 2\delta)(y)$. However, unlike ULA, MALA consists of a second step that will accept the proposal $\wtilde{X}_{n+1}$ with probability
	\[
		\alpha(X_n, \wtilde{X}_{n+1}):= 1\wedge \frac{\pi(\wtilde{X}_{n+1})q(X_n|\wtilde{X}_{n+1})}{\pi(X_n)q(\wtilde{X}_{n+1}|X_n)}.
	\]In other words
	\[
		X_{n+1}=\rl{
			\wtilde{X}_{n+1} & \text{ if }U_n\leq \alpha(X_n, \wtilde{X}_{n+1})\\
			X_{n} & \text{ otherwise}
		};~~~U_n\sim [0,1].
	\]
	\subsection{RMS-Unadjusted Langevin Algorithm} The ULA is a powerful technique for ``black-box'' MCMC in which one only has access to gradient information about the target density. However, for very high dimensional problems in which $\pi$ is highly anisotropic, it can be inefficient to simulate an isotropic diffusion such as the Langevin algorithm. A simple and effective technique, borrowed from the optimization literature in which the same phenomenon can cause problems, is to \emph{precondition} the dynamics as follows. First we maintain an exponentially smoothed squared-gradient estimate $r_n$ defined as
	\[
		r_{n+1} = \beta r_n + (1-\beta)\nabla \log\pi(X_n)^2;~~~\beta\in [0,1[
	\]where the gradient squared is meant element-wise. Then, as in RMSProp \cite{Tieleman2012}, we can construct an adaptive stepsize by dividing by
	the square root of $r_n$ as follows
	\[
		\wat{\delta}_{n+1} := \frac{\delta}{\sqrt{r_{n+1} + \eps}}
	\]and then using this stepsize in the Langevin update
	\begin{align*}
		X_{n+1} &= X_n + \wat{\delta}_{n+1}\nabla \log\pi(X_n) + \sqrt{2\wat{\delta}_{n+1}}Z_n\\
		&= X_n + \frac{\delta}{\sqrt{r_{n+1} + \eps}}\nabla \log\pi(X_n) + \sqrt{2\frac{\delta}{\sqrt{r_{n+1} + \eps}}}Z_n.
	\end{align*}
	This has the effect of using gradient information to scale the stepsize of original Langevin diffusion independently along each dimension, which likely reduces the negative impacts of anisotropy and substantially increases mixing rate. As studied in in \cite{li2016preconditioned}, this algorithm, which we call RMS-ULA, is biased but this bias can be controlled with $\delta$.

	\subsection{RMS-MALA.}
	Similar to the progression from ULA to MALA, we can ``metropolize'' the RMS-ULA algorithm to correct the bias. This follows the same structure as the MALA, except that the proposal depends on $r$\footnote{A ``proper'' setup would expand the state space to include be $(x,r)$, which would restore the Markov property}
	\begin{align*}
		r' &= \beta r + (1-\beta)\nabla \log\pi(x)^2\\
		q(y|x)&=\calN\left(x + \frac{\delta}{\sqrt{r' + \eps}}\nabla \log\pi(X_n), 2\frac{\delta}{\sqrt{r' + \eps}}\right).
	\end{align*}

	\subsection{Practical Considerations for Bayesian Neural Networks: Reversibility \& Tempering.}
	In practical settings, the high dimensional, anisotropic characteristics and limited computation budget of Bayesian neural networks (BNNs) necessitate some modifications to the above algorithms. 

	The first is that the Metropolis-Hastings (MH) step is rarely used except in simple settings. This is because, while ensuring that that $\pi$ is the invariant measure, the reversibility of the MH step is too inefficient when we cannot affort to reject many samples. Because they cannot ``backtrack'', nonreversible dynamics are generally much more efficient than reversible dynamics \cite{hwang2005accelerating,duncan2016variance,pavliotis2014stochastic,barp2022geometric}. This has led to the recent interest in using piecewise deterministic Markov processes \cite{bierkens2019zig,deligiannidis2019exponential,chevallier2021pdmp,durmus2021piecewise} and other nonreversible MCMC dynamics such as \cite{monmarche2021high} for high dimensional Bayesian inference. See \cite{barp2022geometric},~Section 3.1 for an interesting dicussion on nonreversibility and efficiency.

	The second modification is tempering. A tempered version of a distribution $\pi$ is $\pi_\beta\propto \pi^\beta$. This is a pragmatic solution to speed up exploration of a target distribution when the noise component of Langevin-like algorithms causes slow or unstable dynamics. In the limit as $\beta\to \infty$, sampling from $\pi_\infty$ is equivalent to the maximum \emph{a posteriori} estimate (this is simulated annealing \cite{bertsimas1993simulated}). In practice, tempering can be achieved by simply scaling the noise by $\sqrt{\tau}$
	\[
		X_{n+1} = X_n + \delta \nabla \log \pi(X_n) + \sqrt{2\delta}\sqrt{\tau}Z_n,~~~Z_n\sim \calN(0,I_d).
	\]which targets $\pi^\beta$ where $\beta=1/\tau$.\footnote{To see this, for $\pi_\beta\propto\pi^\beta$ we have $\nabla\log\pi_\beta = \beta\nabla\log\pi$ so by using a stepsize $\delta'=\delta/\beta$ we get ${X_{n+1}=X_n + \beta\delta' \nabla\log\pi(X_n) + \sqrt{2\delta'}Z_n=X_n + \delta\nabla\log\pi(X_n) + \sqrt{2\delta/\beta}Z_n}$ and $\tau=1/\beta$.} See also \cite{wenzel2020good} for a discussion of tempering in BNNs.

	\section{Experimental Details}\label{app:exp}
	No hyperparameter sweeps were used in the any of these experiments. Hyperparameters were chosen based on reasonable guesses and minimal manual tuning. 

	\subsection{2d Toy Experiments}\label{app:exp_2d}
	\subsubsection{Maximum Mean Discrepancy}
	We use the maximum mean discrepancy (MMD) metric \cite{gretton2012kernel} as a way to quantitatively evaluate our MCMC algorithms. The MMD metric is an integral probability metric of the form
	\[
		\|\mu - \nu\|_{MMD} = \sup_{\|f\|_{\calH}\leq 1}|\mu(f) - \nu(f)| = \sup_{\|f\|_{\calH}\leq 1}|\E_{X\sim \mu}[f(X)] - \E_{Y\sim \nu}[f(Y)]|
	\]where $\calH$ is a reproducing kernel Hilbert space (RHKS) associated to a positive-definite kernel function $k:\R^d\times \R^d\to \R$; see \cite{gretton2012kernel} for details. For our purposes, the important features of the MMD metric are that 1) it is similar to the total variation metric, which optimizes over the unit ball in $\calB_b(\R^d)$ instead of $\calH$; and 2) the MMD metric $\|\mu-\nu\|_{MMD}$ can be efficiently empirically estimated from samples of $\mu,\nu$. This is because, according to Lemma~6 in \cite{gretton2012kernel}, we have
	\begin{align*}
		\|\mu - \nu\|_{MMD}^2 &= \E_{X,X'\sim \mu\otimes \mu}[k(X,X')] - 2 \E_{X,Y\sim \mu\otimes \nu}[k(X,Y)] + \E_{Y,Y'\sim \nu\otimes \nu}[k(Y,Y')]
	\end{align*}
	which depends only on the kernel function $k$, and an \emph{unbiased} estimator of $\|\mu - \nu\|_{MMD}^2$ is
	\[
		\|\mu - \nu\|_{MMD}^2 \approx \frac{1}{N_\mu(N_\mu-1)}\sum_{i,j=1,~j\neq i}^{N_\mu} k(X^i, X^j) - 2 \frac{1}{N_\mu N_\nu}\sum_{i=1}^{N_\mu}\sum_{j=1}^{N_\nu} k(X^i, Y^j) + \frac{1}{N_\nu(N_\nu-1)}\sum_{i,j=1,~j\neq i}^{N_\nu} k(Y^i, Y^j)
	\]where $X^i\iid \mu$ for $i=1,\dots,N_\mu$ and $Y^i\iid \nu$ for $i=1,\dots,N_\nu$. 

	If we treat $\mu$ as the distribution $\mu_n$ of a (linear or nonlinear) MCMC algorithm at step $n$ and $\nu$ as $\pi$, then we can apply this estimator provided we can sample from $\pi$ exactly. This is not feasible for complex high-dimensional distributions, but it is possible for the toy distributions we have chosen, which are all mixtures or transformations of Gaussians.

	\subsubsection{Experimental Setup}
	Refer to Table~\ref{tab:2d_settings} for the experimental settings. Additionally, we used $10,000$ samples from $\pi$ and the kernel
	\[
		k(x,y) = \exp(-\|x-y\|^2) + \exp(-2\|x-y\|^2)
	\] to estimate the MMD as in the previous section. 

	\begin{table}[ht!]
		\centering
		\caption{Experimental settings for the 2d toy experiments.}
		\label{tab:2d_settings}
		\begin{tabular}{l c C{5cm}}
			\textbf{Setting} & \textbf{Symbol} &  \textbf{Value}\\
			\hline
			Auxiliary/Linear Markov kernel & $Q$ & MALA\\
			Auxiliary/Linear kernel stepsize & $\delta_{aux}$ & $0.001$\\
			Auxiliary/Linear target density & $\eta^\star$ & $\calN(0,\sigma^2I_2)$ with $\sigma=4$ for circular MoG and two rings and $\sigma=20$ for grid MoG\\
			Primary Markov kernel & $K$ & MALA \\
			Primary kernel stepsize & $\delta$ & $0.001$\\
			Number of samples & $N$ & $2000$\\
			Initial Auxiliary/Linear distribution & $\eta_0$ & $\calU([-7.5, 7.5]\times[-7.5, 7.5])$\\
			Initial Primary distribution & $\eta_0$ & $\calU([-7.5, 7.5]\times[-7.5, 7.5])$\\
			Jump probability & $\veps$ & $0.1$ \\
			Number of simulation steps & $n_{\text{sim}}$ & $10,000$\\
		\end{tabular}
	\end{table}

	\subsubsection{Runtime Analysis}\label{app:exp_2d_runtime}
	We also report on the performance vs runtime of the linear and nonlinear algorithms measured in terms of number of gradient executions and the wallclock time. \nnewline

\begin{figure}[ht!]
		\begin{tabular}{C{4.75cm} C{4.75cm} C{4.75cm}}
			\includegraphics[scale=0.175]{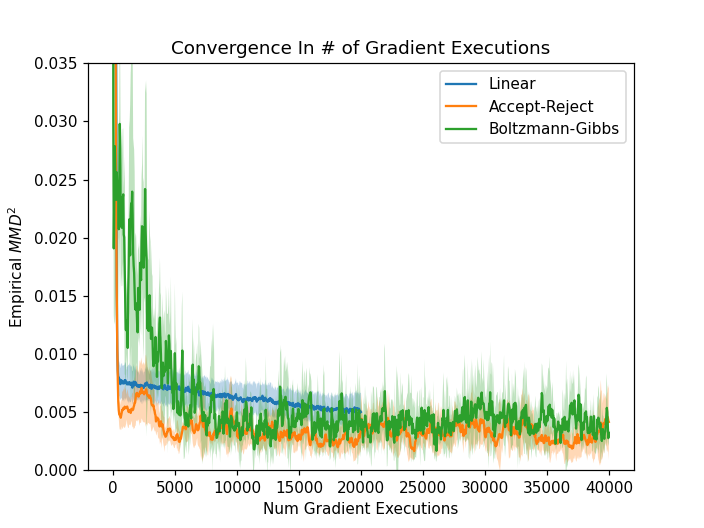}
			& \includegraphics[scale=0.175]{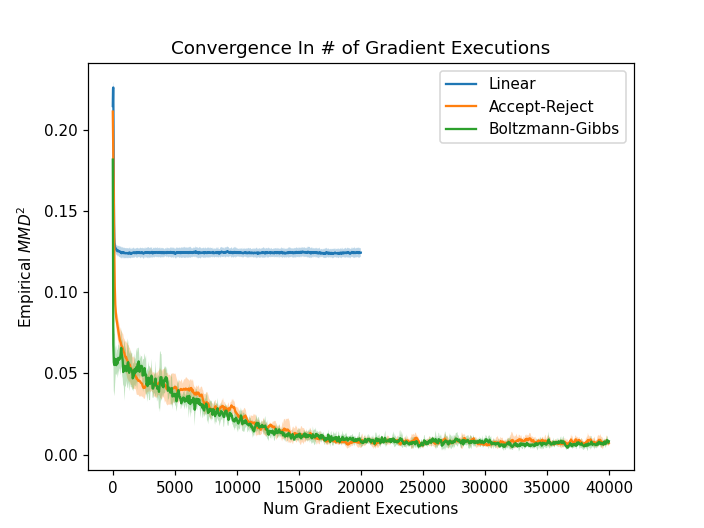}
			& \includegraphics[scale=0.175]{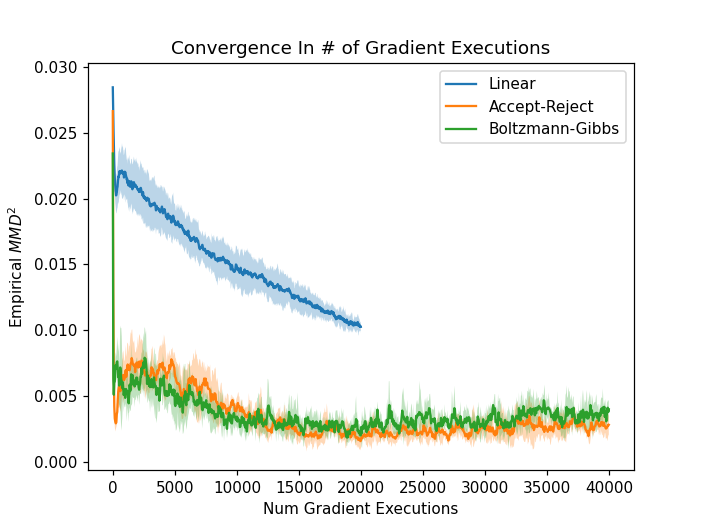} \\
			Circular MoG & Two Rings & Grid MoG
		\end{tabular}
		\caption{$MMD^2$ performance vs number of gradient executions (per sample) for the two-dimensional examples.}
		\label{fig:runtime_2d}
	\end{figure}

	In general, the nonlinear methods will require $2\times$ the number of gradient executions as the linear methods due to the requirement that we sample from the auxiliary chain in addition to the primary chain. In Figure~\ref{fig:runtime_2d}, we show the MMD-squared metric plotted against the number of gradient evaluations. Since we ran all algorithms for a fixed number of steps, this amounts to using $2\times$ the gradient evaluations for the nonlinear methods, although not $2\times$ the steps for a given particle. Despite this, the nonlinear methods offer better performance per gradient evaluation.

	In Table~\ref{tab:2d_walltime}, we also report the wallclock time for each of the methods and each of the distributions. 
	We omit the first twenty iterations in our calculation since these can be slow while JAX compiles the program and we only want to measure the steady-state execution performance.

	\begin{table}[ht!]
		\centering
		\begin{tabular}{l l c c c}
			\textbf{Target Density} & \textbf{Sampler} & \textbf{Wall time (s)} & \textbf{Slowdown vs Linear} \\
			\hline
			\hline
			\multirow{3}{*}{Circular MoG} & Linear & $0.97_{\pm 0.02}$ & -  \\
			& AR & $2.57_{\pm 0.06}$ & 2.64 $\times$ \\
			& BG & $2.99_{\pm 0.01}$ & 3.08 $\times$ \\
			\hline
			\multirow{3}{*}{Two Rings} & Linear & $0.97_{\pm 0.02}$ & - \\
			& AR & $2.57_{\pm 0.06}$ & 2.65 $\times$ \\
			& BG & $3.18_{\pm 0.01}$ & 3.27 $\times$ \\
			\hline
			\multirow{3}{*}{Grid MoG} & Linear & $1.33_{\pm 0.01}$ & - \\
			& AR & $2.66_{\pm 0.01}$ & 2.00 $\times$ \\
			& BG & $3.70_{\pm 0.01}$ & 2.79 $\times$ \\
			\hline
		\end{tabular}
		\vspace{2mm}
		\caption{Wall Time analysis of the two-dimensional sampling problems (mean $\pm 1$ standard deviation on $5$ trials). We measure the steady-state wall time by omitting the first 20 iterations to account for the JAX jit-compilation time at the start of the program.}
		\label{tab:2d_walltime}
	\end{table}

	\subsubsection{Comparison With Algorithms from \cite{delmoralnonlin}}\label{app:dm_comparison}
	Thanks to a very helpful suggestion from an Anonymous Reviewer, we conducted a comparison of the nonlinear MCMC algorithms proposed in this paper with those proposed in the ``parent'' work of this paper, i.e. \cite{delmoralnonlin}. Below, we detail the implementation details as they are nontrivial, and the results we obtained. See also Figure~\ref{fig:us_vs_delmoral} for the visual results.

	\paragraph{Implementation Details.}
	As detailed in Appendix~\ref{app:diffs}, the key difference between algorithms in this work and those in \cite{delmoralnonlin} is how one constructs the empirical measure $\eta^N_n$ for indexing $K_{\eta^N_n}$ in \eqref{eq:general_IPS}. Recall, ours uses a fixed batch of samples, and the algorithm from \cite{delmoralnonlin} uses all past samples from a single trajectory. In other words, our approach uses $\eta^N_n = m(\{Y^1_n, \dots, Y^N_n\})$ whereas \cite{delmoralnonlin} uses $\eta^n_n = m(\{Y_0, \dots,Y_n\})$.

	This decision to use a dynamic number of samples is well-motivated from a theoretical point of view in \cite{delmoralnonlin}, but is highly suboptimal from an implementation standpoint. The reason is that using a dynamic number of samples, i.e. $n$ in the case of \cite{delmoralnonlin}, results in frequent memory reallocations which are costly operations in common software frameworks and especially on GPUs. Indeed, modern software frameworks and accelerators have extensive optimizations for fixed-size (i.e. batched) workloads and this approach is contrary to this paradigm. In fact, this work was motivated by the need to understand how fixed sample sizes would perform in the setting introduced by \cite{delmoralnonlin}.

	Nevertheless, to implement the algorithm in \cite{delmoralnonlin} as a benchmark, we were faced with the classical memory-speed trade-off in software engineering. On the one hand, one can disable jit-compilation in \verb|JAX| and accept the memory-allocation costs to use a variable-size array to store the samples. Alternately, one can aggressively use more memory to speed up the computation by pre-allocating enough device memory for all future samples and apply masking to do computations on a subset of those samples. The former approach yielded a slowdown of $\sim 140\times$ on the 2-dimensional examples, which was deemed unacceptable. Hence we used the pre-allocation method at the cost of allocating a $N\times n\times d$-\verb|float32| tensor. For small $d$, i.e. 2 in our case, this was acceptable at $2,000\times 10,000\times 2\times 4=$152MB. However, for larger-scale problems such as the CIFAR10 experiments below, this would be completely intractable with either approach.

	\paragraph{Bias-Implementation Trade-off.}
	The results of our comparison experiment unequivocally support the bias-speed trade-off we expect in our work. From Figure~\ref{fig:us_vs_delmoral}, we see that the algorithms from \cite{delmoralnonlin} are considerably more stable and have lower variance \emph{and} bias. This is explainable by comparing the effective number of samples each method uses -- ours uses $N=2,000$ throughout, but theirs effectively uses $N=10,000$ at the end of simulation. In our main result Theorem~\ref{thm:main_result}, we show that the bias of our algorithm decays like $1/N$ for fixed timestep $n$, therefore using more samples in our method would result in a less-biased approximation of $\pi$. On the other hand, as shown in \cite{delmoralnonlin}, their algorithm is asymptotically \emph{unbiased}. Despite this desirable property, the discussion above clearly shows why accepting \emph{some} bias is required for practical applications of this type of algorithm.

	In Figure~\ref{fig:2d_exp_10k}, we show the results of a study  matching the $N=10,000$ samples used by the algorithms from \cite{delmoralnonlin} in our own algorithms. The results align completely with the predictions made by Theorem~\ref{thm:main_result}; i.e. the bias of our methods decreases significantly. We see it is also comparable to that of the methods in \cite{delmoralnonlin} for many of the samplers and distributions used in these experiments. We note that this study is a qualitative example of the fundamental difference between nonlinear and linear MCMC algorithms: increasing $N$ directly impacts the convergence rate of our algorithms but has \emph{no effect} on the convergence of the linear ones. This is due to the interacting (i.e. nonlinear) nature of our MCMC methods, and is supported by all the theoretical results we have developed so far.

	\paragraph{Convergence Rate.}
	From Figure~\ref{fig:us_vs_delmoral}, we see empirically that our algorithms have better convergence rates than those in \cite{delmoralnonlin}. While the results in \cite{delmoralnonlin} are asymptotic and hence make no claim about rate of convergence, this aligns with our intuition of how these samplers work. In particular, using past samples --- including from the very beginning of simulation --- can slow down the convergence since one expects that \emph{future} samples will be of higher quality than \emph{past} samples for a \emph{converging} algorithm. In \cite{delmoralnonlin}, they use a burn-in period but of course this does nothing to help with convergence rate. On the other hand, Theorem~\ref{thm:main_result} upper-bounds convergence by the ``pseudo-geometric'' rate $\calO(n\rho^n)$ for $\rho<1$.

	\begin{figure}[ht!]
		\centering
		\begin{tabular}{C{3.5cm} C{3.5cm} C{3.5cm}}
			\includegraphics[scale=0.175]{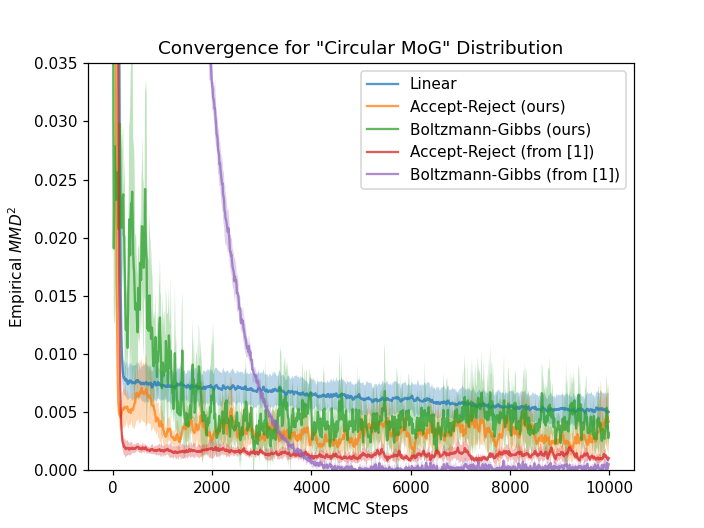} & \includegraphics[scale=0.175]{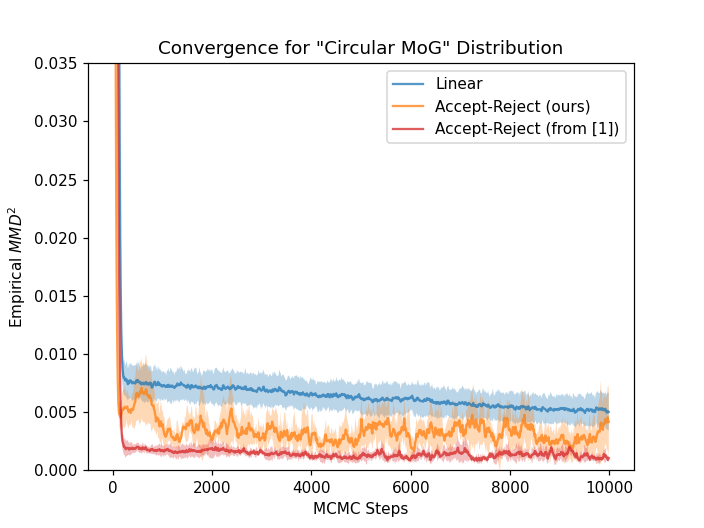}
			& \includegraphics[scale=0.175]{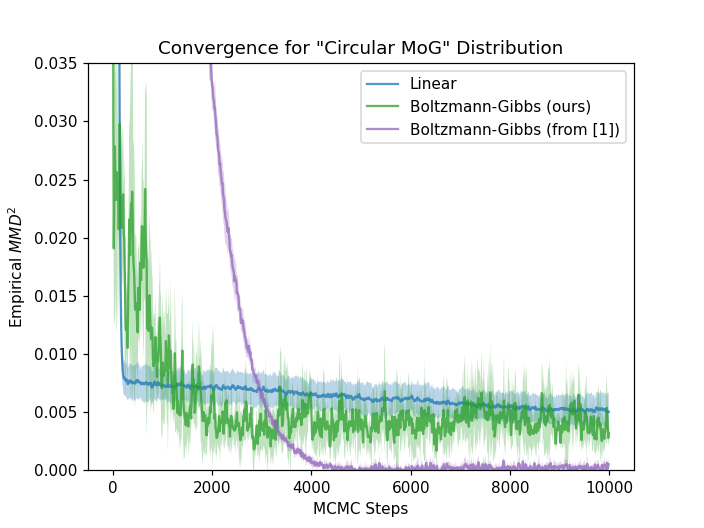} \\
			\includegraphics[scale=0.175]{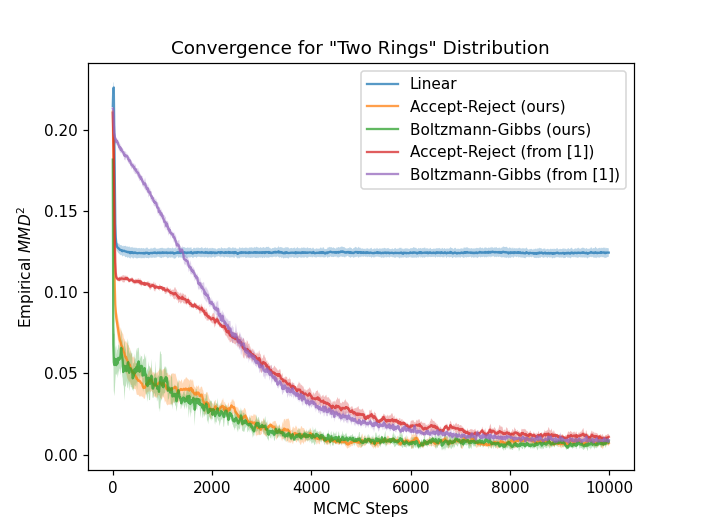} & \includegraphics[scale=0.175]{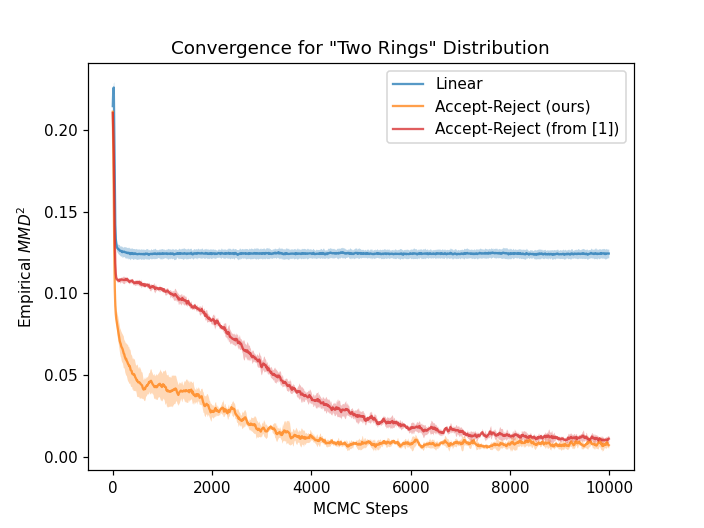}
			& \includegraphics[scale=0.175]{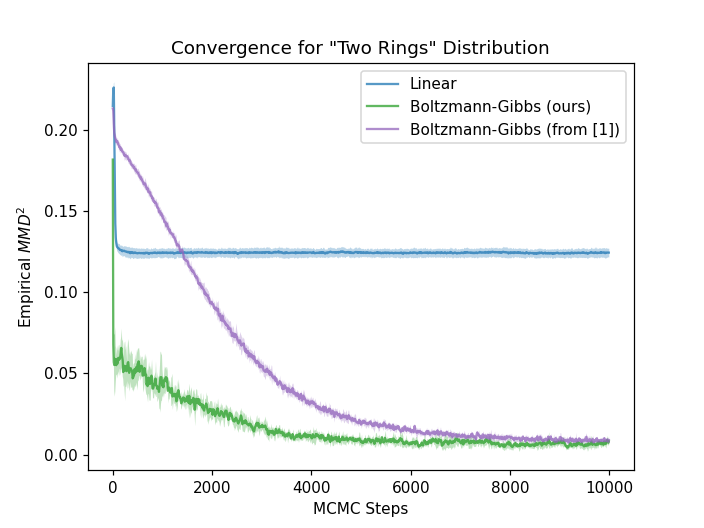} \\
			\includegraphics[scale=0.175]{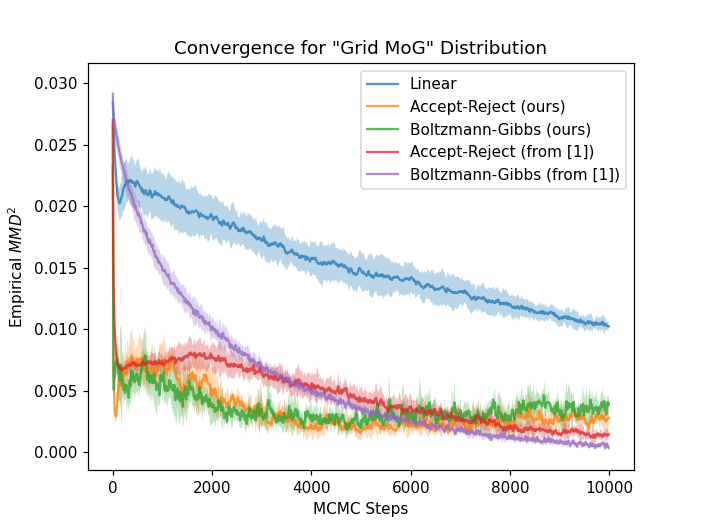} & \includegraphics[scale=0.175]{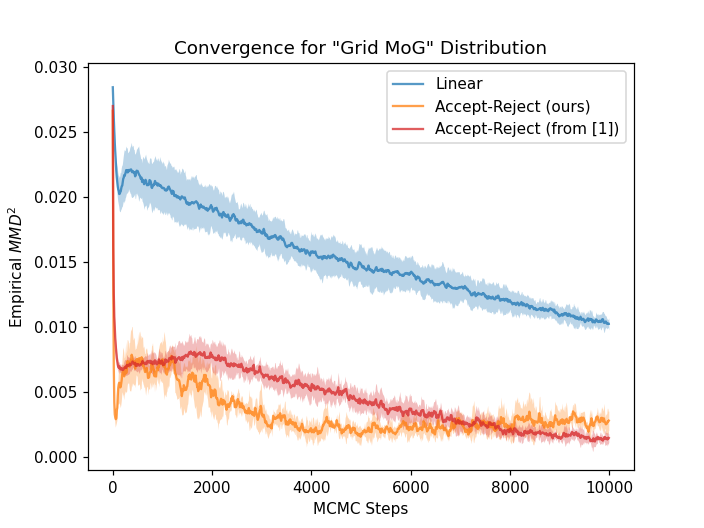}
			& \includegraphics[scale=0.175]{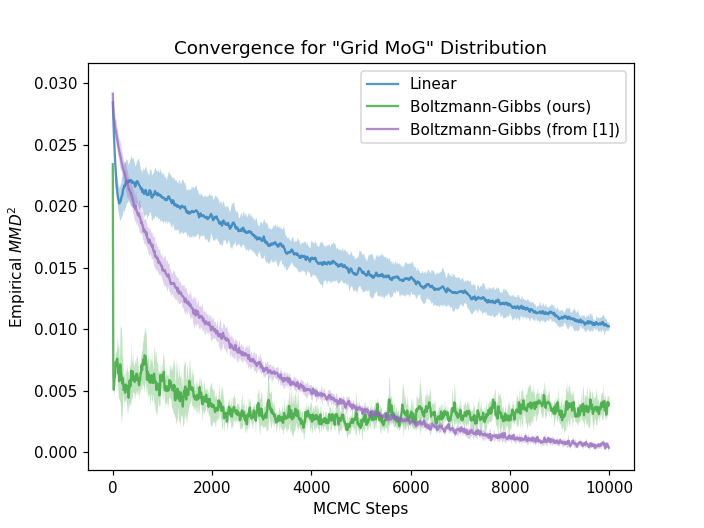}\\
		\hspace{6mm} All Algorithms & \hspace{8mm} Accept-Reject & \hspace{6mm} Boltzmann-Gibbs
		\end{tabular}
		\caption{A comparison of the empirical MMD-squared performance for our nonlinear MCMC algorithms with $N=2,000$ particles against those of \cite{delmoralnonlin} in the same setup as Section~\ref{subsec:2d}. The rows show the convergence for the Circular Mixture of Gaussians (MoG) density \cite{stimper2021resampling}, the Two Rings density \cite{stimper2021resampling}, and the Grid Mixture of Gaussians density \cite{zhang2019cyclical} respectively.}
		\label{fig:us_vs_delmoral}
	\end{figure}

	\begin{figure}[ht!]
		\centering
		\begin{tabular}{C{3.5cm} C{3.5cm}}
			\includegraphics[scale=0.175]{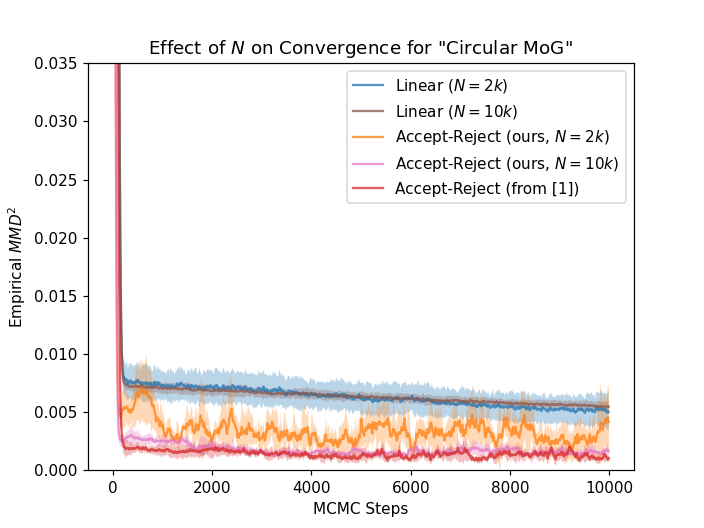}
			& \includegraphics[scale=0.175]{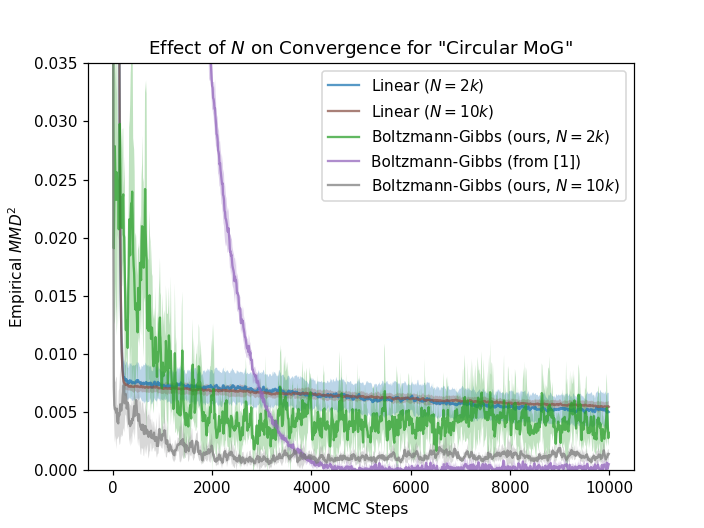} \\
			\includegraphics[scale=0.175]{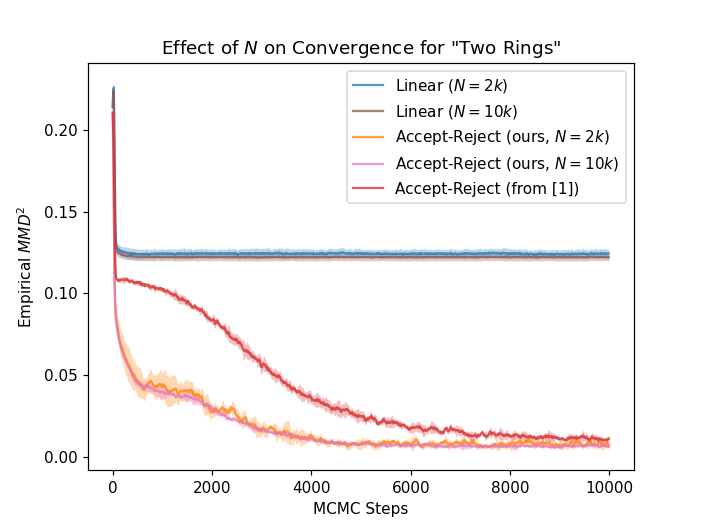}
			& \includegraphics[scale=0.175]{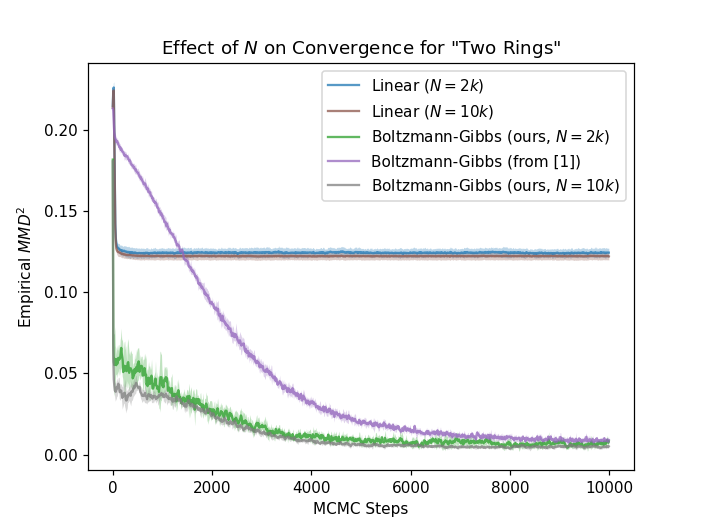} \\
			\includegraphics[scale=0.175]{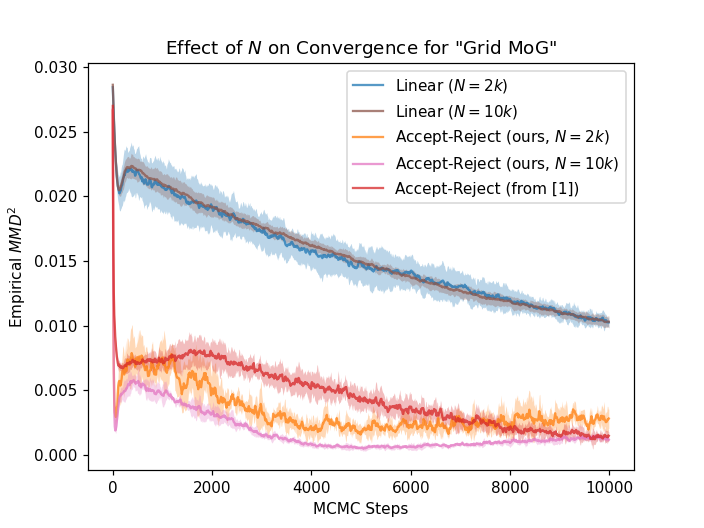}
			& \includegraphics[scale=0.175]{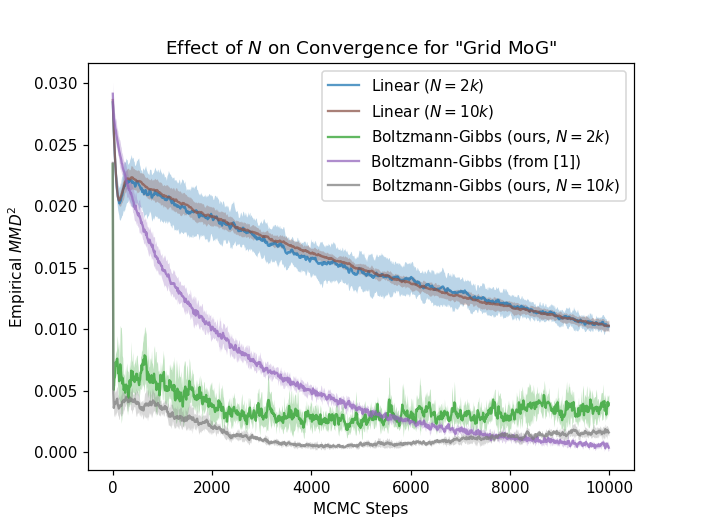} \\
			\hspace{8mm} Accept-Reject & \hspace{6mm} Boltzmann-Gibbs
		\end{tabular}
		\caption{An illustration of the effect of increasing the number of particles $N$ on our nonlinear algorithms. We show how increasing $N$ from $N=2,000$ in previous experiments (e.g. Figure~\ref{fig:us_vs_delmoral}) to $N=10,000$ affects the rate of convergence and bias for our nonlinear algorithms. $N=10,000$ was chosen purposefully to match the effective number of particles used in $\eta^N_n$ for the algorithms in \cite{delmoralnonlin} to provide a fair comparison of bias.}
		\label{fig:2d_exp_10k}
	\end{figure}

	\subsection{CIFAR10 Experiments}\label{app:exp_cifar}

	\subsubsection{Experimental Setup}
	In this experiment, we implement a Bayesian neural network for image classification on the CIFAR10 dataset \cite{krizhevsky2009learning}. The neural network architecture we use is a standard ResNet-18 architecture \cite{he2016deep} using the default settings implemented by the DeepMind \verb|haiku| library \cite{haiku2020github}. The likelihood function $P(y|x,\theta)$ is the standard crossentropy on the logits $y$ given an input image $x$ and parameters $\theta$. We also use standard data augmentation techniques including standardizing the images by the mean and variance, random crops and random flips. See the Table~\ref{tab:cifar_settings} for the experimental parameters. 

	\begin{table}[ht!]
		\centering
		\caption{Experimental settings for the CIFAR10 experiments.}
		\label{tab:cifar_settings}
		\begin{tabular}{l c C{5cm}}
			\textbf{Setting} & \textbf{Symbol} &  \textbf{Value}\\
			\hline
			Auxiliary Markov kernel & $Q$ & RMS-Langevin\\
			Auxiliary kernel stepsize & $\delta_{aux}(n)$ & $0.001 \times 0.1^{\lfloor n/2000\rfloor}$\\
			Auxiliary RMS $\beta,\eps$ parameters & N/A & $0.9,1\times 10^{-9}$\\
			Auxiliary target density & $\eta^\star$ & $\eta^\star\propto \pi^{0.9/\tau}$\\
			Noise scaling (tempering) &  $\sqrt{\tau}$ & $1\times 10^{-4}$ \\
			Primary Markov kernel & $K$ & Unadjusted Langevin \\
			Primary kernel stepsize & $\delta$ & $5\times 10^{-5}$\\
			Number of samples & $N$ & $10$\\
			Minibatch size & $|\wat{\calD}|$ & 256 \\
			Initial Auxiliary distribution & $\eta_0$ & $\calN(\theta_0, 0.001)$ where $\theta_0$ is the initialized set of parameters from the \verb|Haiku| implementation\\
			Initial Primary distribution & $\mu_0 $ & same as $\eta_0$\\
			Jump probability & $\veps$ & $0.05$ \\
			Number of simulation steps & $n_{\text{sim}}$ & $50$ passes through the dataset \\
			Bayesian Prior & $P(\theta)$ & $\calN(0,1\times 10^{-4} I_d)$ \\
		\end{tabular}
	\end{table}

	\subsubsection{Additional Plots}\label{app:cifar10_addtl_plots}
\begin{figure}[ht!]
		\centering
		\includegraphics[scale=0.3]{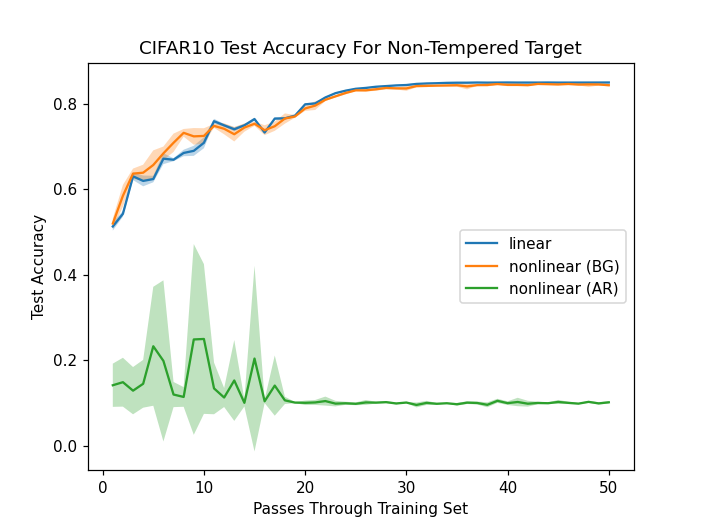}
		\caption{The full set of curves for the non-tempered CIFAR10 experiment. The AR sampler diverged early during sampling and was not able to achieve better than random performance. This shows that tempering is a useful technique for improving stability of these high-dimensional Bayesian sampling problems.}
		\label{fig:diverged}
	\end{figure}

	See Figure~\ref{fig:diverged} for the plot of CIFAR10 test accuracy during sampling for the non-tempered case. 

	\subsubsection{Calibration Analysis}
	Calibration of a classifier is a measure of how well that classifier's logits represent probabilities \cite{guo2017calibration}. We use this as a necessary but not sufficient test for correct sampling, since we know \emph{a priori} that the Bayesian posterior gives true probabilities. We will use the methodology in \cite{guo2017calibration}, particularly ``expected calibration error'' and ``maximum calibration error'' to study the calibration of our Bayesian neural network, which we describe below.

	To measure calibration, assume first that we have a test set of data $\calD_{\text{test}}=\{(x_1,y_1),\dots,(x_{N_{\text{test}}}, y_{N_{\text{test}}})\}\subset \calX\times \{0,\dots,C\}$ where $\calX$ is the input space and there are $C\in \Z_{>1}$ labels. Suppose $\wat{p}:\calX\to \Delta^{C-1}$ is a supposed probability distribution representing $p(y|x)$. If $\wat{p}$ were a good representation of $p(y|x)$, we would expect that the values of $\wat{p}$ should correlate with the accuracy of $\wat{p}$, i.e. when $\wat{p}$ is $\alpha$\% certain, then $\alpha$\% of the time $\wat{p}$ is correct. More formally, perfect calibration is when
	\[
		P(y=y_i | \wat{p}[y_i]=\alpha)=\alpha.
	\]

	We can empirically estimate this relationship between predicted probability and true probability using histograms. Specifically, given predicted probabilities $\wat{p}_i=\wat{p}(y_i|x)$, we can form bins $B_k,~k=1,\dots,M$ on $[0,1]$ and estimate the accuracy in the $k$-th bin as $A_k$ and the average probability in that bin as $\alpha_k$. If $\wat{p}$ represents a true probability distribution, then $A_k\approx \alpha_k$ for each $k=1,\dots,M$. The expected calibration error (ECE) is then
	\[
		ECE(\wat{p}):= \sum_{k=1}^M \frac{|B_k|}{N_{\text{test}}}|A_k - \alpha_k|.
	\]We can also study the maximum calibration error
	\[
		MCE(\wat{p}) := \max_{k=1,\dots,M}|A_k- \alpha_k|.
	\]

	In Table~\ref{tab:cifar_calibration}, we report the ECE and MCE for the CIFAR10 experiment, where now
	\[
		\wat{p}(y|x, \calD) = \sum_{i=1}^N p(y|x,\theta^i) \approx \int p(y|x,\theta)p(\dd\theta|\calD).
	\]

	\begin{table}[ht!]
		\caption{Calibration errors for the CIFAR10 experiment. $5$ runs were used, and $\pm$ represents one standard deviation. Both ECEs and tempered MCE appear to be statistically significantly lower for the nonlinear than the linear algorithm, indicating better calibration in those cases. The ECE numbers have been multiplied by $10^2$ in this table.}
		\label{tab:cifar_calibration}
		\vspace{2mm}
		\begin{tabular}{l c c c c c}
			& \multicolumn{2}{c}{\textbf{Expected Calibration Error} ($\downarrow$)} & \multicolumn{2}{c}{\textbf{Max Calibration Error} ($\downarrow$)}\\
			Algorithm & Non-Tempered & Tempered & Non-Tempered & Tempered\\
			\hline
			Linear & $0.24_{\pm0.02}$ & $0.26_{\pm0.014}$ & $3.78_{\pm0.43}$ & $4.88_{\pm0.85}$\\ 
			\hline
			Nonlinear (BG) & $0.14_{\pm0.03}$ & $0.16_{\pm0.03}$ & $3.21_{\pm0.75}$ & $3.72_{\pm0.40}$\\
			$p$-values BG vs Linear & \textbf{0.00063} & \textbf{0.00028} & 0.22143 & \textbf{0.03953} \\
			\hline
			Nonlinear (AR) & \emph{Diverged} & $0.15_{\pm0.05}$ & \emph{Diverged} & $3.45_{\pm 1.09}$\\
			$p$-values AR vs Linear & - & \textbf{0.000748} & - & 0.28993 \\
			\hline
		\end{tabular}
	\end{table}

	\subsubsection{Distribution Shift}\label{app:exp_cifar_ood}
	One useful outcome of using the probabilistic paradigm for ML is robustness to distribution shift. This is well-documented \cite{gal2016dropout,guo2017calibration}; traditional ML models tend to be overconfident even in settings where they have no hope of making good predictions. In short, traditional ML models lack the ability to say ``I don't know'' in a way that probabilistic ML models do not. This is particularly salient in real-world systems such as self-driving cars, where the distribution of images is highly nonstationary and the costs for overconfident predictions are high.

	In Figure~\ref{fig:cifar_ood} below, we show how the predictive entropy differs between in-domain examples, i.e. the CIFAR10 test set, and out-of-domain examples. For the out-of-domain examples, we use the SVHN dataset \cite{netzer2011reading} which also contains $32\times32$ RGB images, but this time of housing numbers (i.e. the digits 0-9) rather than objects such as airplanes or dogs. The entropy (a measure of uncertainty) is calculated as
	\[
		H(\mbf{p}) = -\sum_{i=1}^{10} p_i \log p_i
	\]where $\mbf{p}\in \Delta^9$ is the vector of probabilities for a prediction $p(y|x,\calD_{\text{train}})$. In the traditional ML case, the probabilities are made with $p(y|x,\theta^*)$. We average $H(\mbf{p})$ over the test data for the in-domain and out-of-domain examples. The optimized method was trained to the same test accuracy ($85\%$) using essentially the same hyperparameters (learning rate, etc). 

	\begin{figure}
		\begin{tabular}{c c c}
			\includegraphics[scale=0.2]{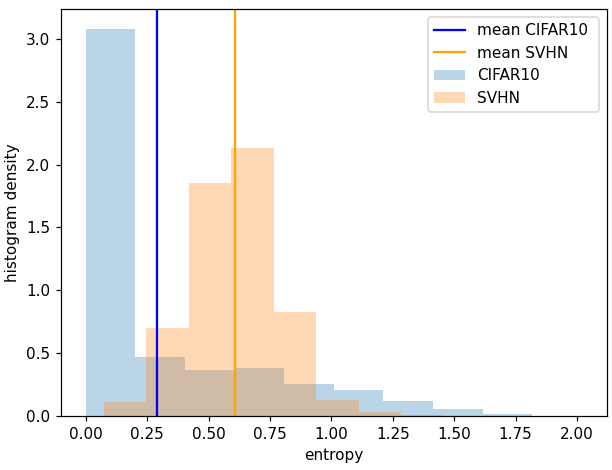} 
			& \includegraphics[scale=0.2]{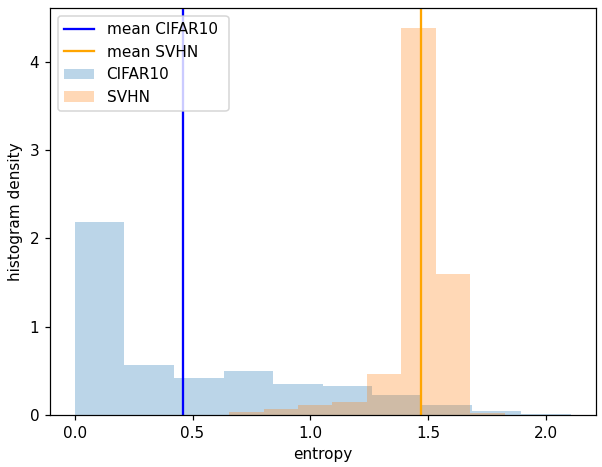} 
			& \includegraphics[scale=0.2]{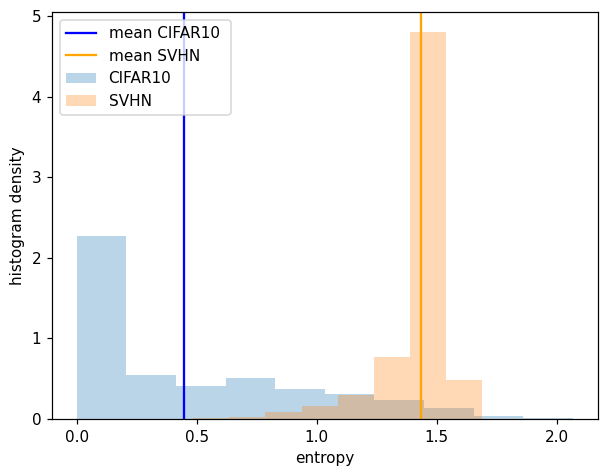} \\
			Optimized & Linear & Nonlinear (BG)
		\end{tabular}
		\caption{Demonstration of the in/out of domain performance of optimized, linear MCMC, and nonlinear MCMC. The linear and nonlinear methods perform essentially the same, which is \emph{expected} since they both approximate the same distribution $P(y|x,\calD_{\text{train}})$}
		\label{fig:cifar_ood}
	\end{figure}

	\subsubsection{Runtime Analysis}\label{app:cifar10_runtime}
	In Figure~\ref{app:cifar10_runtime}, we plot the performance of the tempered target vs the number of gradient evaluations. As described in Appendix~\ref{app:exp_2d_runtime}, the nonlinear algorithm uses $2\times$ the gradient computations. When plotting the eval performance vs the number of computations, we see that the linear algorithm is significantly more efficient. Further work into designing nonlinear samplers that achieve better performance (such as in the two-dimensional examples above) is required to make this trade-off worthwhile in practice.

	\begin{figure}[ht!]
		\centering
		\includegraphics[scale=0.4]{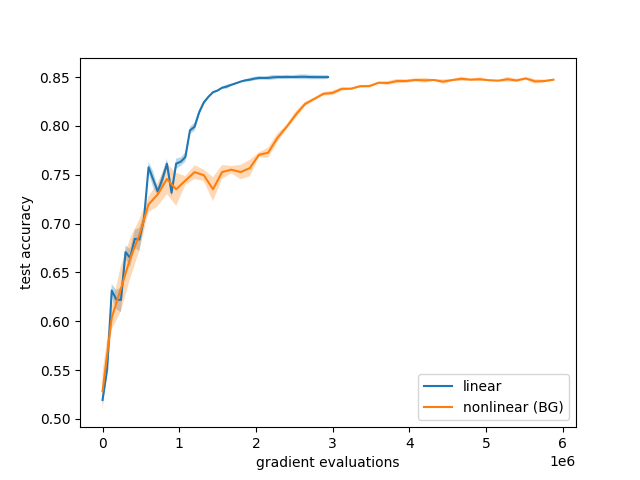}
		\caption{CIFAR10 test performance plotted against number of gradient evaluations (per sample) on the tempered target density, for the linear and nonlinear (BG) MCMC algorithms.}
		\label{fig:cifar10_runtime}
	\end{figure}

	\section{Notation \& Assumptions}\label{app:notation}
	The remaining sections will be devoted to proving the theoretical results contained in this paper. They should be read in sequence. 

	Note that, while some of the assumptions found below appear in \cite{delmoralnonlin}, our analysis and the results we obtain are new. In particular, \cite{delmoralnonlin} studies a different MCMC algorithm based on $K_\eta$, see Appendix~\ref{app:code} for a discussion of algorithm differences. Moreover, they obtain an asymptotic ``strong law of large numbers'' result whereas we obtain nonasymptotic mean field long-time convergence and uniform propagation of chaos results, which are qualitatively different results using different proof techniques.
	\subsection{Notation and Definitions}
	\subsubsection{Probability Spaces, Measures, and Kernels}
	We will be working on the measurable space $(\R^d, \scrB(\R^d))$ with $\scrB(\R^d)$ the Borel $\sigma$-algebra\footnote{This choice is merely for simplicity of exposition; most if not all our results will hold for any polish space and its Borel $\sigma$-algebra.}; let $\calP(\R^d)$ denote the space of probability measures on $\scrB(\R^d)$. Let $\mu\in \calP(\R^d)$, $f:\R^d\to \R$ be a measurable function, and, $K:\R^d\times \scrB(\R^d)\to [0,1]$ be a Markov kernel. Throughout the sequel, we will write
	\[
		\mu(f):=\int f(x)\mu(\dd x), ~~~ Kf(x) := \int f(y) K(x,\dd y),~~~\mu K(\dd y) := \int \mu(\dd x)K(x,\dd y).
	\]An obvious consequence of this is the notation $\mu K(f)=\int f(y)\mu(\dd x)K(x,\dd y)$. We will also need the following subset of $\calP(\R^d)$ for our results: let $G,U:\R^d\to [0,\infty[$ and fix constants $m_G, M_U>0$; then we define
	\[
		\calP_{m_G, M_U}(\R^d):=\{\mu\in \calP(\R^d)\st \mu(G)>m_G,~\mu(U)\leq M_U\}.
	\]We will abbreviate $\calP_{m_G, M_U}(\R^d)=\calP_{m, M}(\R^d)$ when there is no chance of confusion.\nnewline

	We will also frequently use the following notation for the empirical measure associated to $\ovl{y}:=\{y^1,\dots,y^N\}\subset\R^d$:
	\[
		m(\ovl{y}) := \frac{1}{N}\sum_{i=1}^N \delta_{y^i}
	\]with $\delta_x\in \calP(\R^d)$ the Dirac measure. Additionally, we will need the notion of tensor products for functions and measures: let $q\in \N$, $f_i:\R^d\to \R$ and $\mu_i\in \calP(\R^d)$ for $i=1,\dots,q$. Then for $x=(x^1,\dots,x^q)\in (\R^d)^{q}$ and measurable $g:(\R^d)^q\to \R$
	\[
		f_1\otimes \cdots\otimes f_q(x):=f_1(x^1)\cdots f_q(x^q),~~\mu_1\otimes \cdots\otimes \mu_q(g):=\int g(x^1,\dots,x^q)\mu_1(\dd x^1)\cdots\mu_q(\dd x^q).
	\]When $f=f_1=\cdots=f_q$ and $\mu=\mu_1=\cdots=\mu_q$, we will write $f^{\otimes q}$ and $\mu^{\otimes q}$ respectively. Finally, we write $\mu\ll \nu$ to mean that $\nu$ dominates $\mu$ and we write $\diff{\mu}{\nu}$ for the Radon Nikodym derivative. The notation $\mu\sim \nu$ means that $\mu\ll \nu$ and $\nu\ll \mu$.

	\subsubsection{Norms}
	Denote by $\calB_b(\R^d)$ the set of bounded measurable functions $f:\R^d\to \R$. We will be working with a family of norms on functions, and its dual norms on probability measures, which is parameterized by functions of the form $U:\R^d\to [1,\infty[$. For such a $U$, we define 
	\[
		\|f\|_U:=\sup_{x\in \R^d}\frac{|f(x)|}{U(x)}
	\]for $f:\R^d\to \R$. The dual norm to $\|f\|_U$ corresponds to the weighted total variation distance; for $\mu,\nu\in \calP(\R^d)$ we have
	\[
		\|\mu-\nu\|_{tv, U}:=\sup_{\|f\|_{U}\leq 1}|\mu(f) - \nu(f)|.
	\]Note that we could replace the condition $\|f\|_U\leq 1$ by $|f|\leq U$. For $V:\R^d\to [0,\infty[$, we will often work with $V_\beta(x):=1 + \beta V(x)$ for $\beta>0$; in this case, we will write $\|f\|_\beta:=\|f\|_{V_\beta}$ and $\|\bdot\|_{tv,\beta}:=\|\bdot\|_{tv,V_\beta}$. We will also need the definition of the maximum oscillation of $f:\R^d\to \R$, which we denote and define as $\osc(f):=\sup\{|f(x) - f(y)|\st x,y\in \R^d\}$.\nnewline

	For Markov kernels $K,Q$ on $\R^d$, we obtain the weighted kernel distance
	\[
		\|K - Q\|_{ker,U} := \sup_x \frac{\|K(x,\bdot) - Q(x,\bdot)\|_{tv,U}}{U(x)}.
	\]It is worth noting the special case $U\equiv \mbf{1}$ which corresponds to the usual total variation distance; in this case we will write $\|f\|_\infty$ and $\|\mu-\nu\|_{tv}$ in place of $\|f\|_U$ and $\|\mu-\nu\|_{tv,U}$ respectively. Lastly, we will use the notation $\eps(K)$ to denote the contraction coefficient of a Markov kernel $K$ defined as $\eps(K) := \inf\{c \in [0,1] \st \|\mu K - \nu K\|_{tv}\leq c\|\mu - \nu\|_{tv}~\forall \mu,\nu\in \calP\}$ see, e.g. \cite{del2004feynman} Ch 4.
	
	\subsection{Assumptions}
	\begin{assumption}[Drift and Minorization]\label{assump:FL}
		Let $K,Q:\R^d\times \scrB(\R^d)\to [0,1]$ be Markov kernels. 
		\begin{description}
			\item[K1] $K$ satisfies the drift criterion
			\[
				KV(x) \leq aV(x) + b
			\]for $a\in ]0,1[$, $b>0$, and $V:\R^d\to [0,\infty[$ with $\lim_{\|x\|\to \infty}V(x)=\infty$.
			\item[K2] $K$ satisfies the following uniform minorization condition on the level sets of $V$: there exists $\ovl{\gamma}\in ]0,1[$, $\nu\in \calP(\R^d)$, and $R>2b/(1-a)$ s.t.
			\[
				\inf_{\{x\st V(x)\leq R\}}K(x,A)\geq \ovl{\gamma}\nu(A)~~~\forall A\in \scrB(\R^d).
			\]
			\item[Q1] $Q$ satisfies the drift criterion
			\[
				QU(x) \leq  \xi U(x) + c
			\]for $\xi\in ]0,1[$, $c>0$, and $U:\R^d\to [1,\infty[$ with $\lim_{\|x\|\to \infty}U(x)=\infty$. 
			\item[Q2] $Q$ satisfies the following uniform minorization condition on the level sets of $U$: there exists $\zeta\in ]0,1[$, $\nu'\in \calP(\R^d)$, $R'>2c/(1-\xi)$ s.t.
			\[
				\inf_{\{x\st U(x)\leq R'\}}Q(x,A)\geq \zeta\nu'(A)~~~\forall A\in \scrB(\R^d).
			\]
		\end{description}
	\end{assumption}
	The requirements in Assumption~\ref{assump:FL} are from \cite{hairer2011yet} and are by now fairly standard in the Markov chain literature. They will in particular imply the following properties for $K$ and $Q$ respectively:
	\begin{proposition}[Basic Properties of $K,Q$; \citep{hairer2011yet}\&\citep{delmoralnonlin}]
		~
		\begin{enumerate}
			\item Suppose Assumptions~\ref{assump:FL}-K1,K2 hold. Then there exists $\gamma\in ]0,1[$ and $\beta>0$ s.t. $\forall \mu,\nu\in \calP(\R^d)$, we have
			\[
				\|\mu K - \nu K\|_{tv,\beta}\leq \gamma\|\mu - \nu\|_{tv,\beta}.
			\]
			\item Suppose that Assumptions~\ref{assump:FL}-Q1,Q2 hold. Let $\eta_n = \eta_0 Q^n$ for $\eta_0\in \calP(\R^d)$, and $r\in ]0,1]$. If $\eta_0(U^{r})<\infty$, then for $r\in ]0,1[$ there are constants $M(r)>0,~\delta\in ]0,1[$ s.t.
			\begin{equation}\label{eq:delta_defn}
				\|\eta_n - \eta^\star\|_{tv,U^r}\leq M(r)\delta^n.
			\end{equation}
		\end{enumerate}
	\end{proposition}
	\begin{proof}
		\hfill
		\begin{enumerate}
			\item This is directly from \cite{hairer2011yet}.
			\item This uses Lemma~\ref{lemma:conversion} which shows Assumption~\ref{assump:FL}~Q1,Q2 implies the assumptions in Lemma~C.1 from \cite{delmoralnonlin}. Then we combine this with Lemma~\ref{lem:v_tv}. Both lemmas can be found in Appendix~\ref{app:longtime}.
		\end{enumerate} 
	\end{proof}
	Throughout the remainder of this paper, $\beta$ and $\gamma$ will be fixed and we will adopt the notation 
	\[
		V_\beta(x):=1 + \beta V(x).
	\] Additionally, references to $M(r)$ and $\delta$ are meant in the sense of the above proposition. Next, we introduce some criteria that ensure that the kernels $K,Q$ are ``compatible'' in-terms of their drift criteria from Assumption~\ref{assump:FL}.
	\begin{assumption}[Compatibility]\label{assump:compat}
		\hfill
		\begin{description}
			\item[C1] There is $r^*\in ]0,1]$ s.t. $V_\beta(x)\leq U(x)^{r^*}~\forall x\in \R^d$ with $V_\beta$ as above.
			\item[C2] $G$ satisfies the lower bound compatibility criterion with $U$: for every $R>0$
			\[
				\theta(R):=\inf\{G(x) \st x\in \R^d,~U(x)\leq R\} > 0.
			\]
		\end{description}
	\end{assumption}
	Assumption~\ref{assump:compat}-C1 is also present in \cite{delmoralnonlin}, and ensures that $V$ and $U$ are sufficiently ``comparable''. Assumption \ref{assump:compat}-C2 is a novel assumption that we will use to obtain an \emph{a priori} lower bound on $\eta_n(G)$. Finally, some straightforward boundedness assumptions on $G$.

	\begin{assumption}[Assumptions on $G$]\label{assump:G}
		\hfill
		\begin{description}
			\item[G1] $G$ is bounded, i.e. $\|G\|_\infty<\infty$.
			\item[G2] $G$ is bounded in the weighted supremum norm for $V_\beta$; i.e. $\|G\|_\beta<\infty$. 
		\end{description}
	\end{assumption}
	Not all of the results below depend on all of the assumptions in this section. We will make clear which of these assumptions are invoked in each result.

	\section{Long-Time Convergence}\label{app:longtime}
	We will now state the main long-time convergence result for the mean field system. We will revisit this theorem at the end of this section and provide a full proof. We will often use the equivalent ``distribution flow'' interpretation of system \eqref{eq:gen_mf_system} defined as
	\begin{equation}\label{eq:gen_mf_flow}
		\rl{
			\eta_{n+1}&= \eta_n Q\\
			\mu_{n+1}&=\mu_n K_{\eta_{n+1}}
		}
	\end{equation}
	for our proofs.

	\begin{restatable}{theorem}{genconv}\label{thm:gen_conv}
		Suppose that Assumption~\ref{assump:FL} and Assumption~\ref{assump:compat}-C1 hold. Suppose also that $J_\eta$ satisfies
		\begin{equation}\label{eq:gen_reg}
			\|J_{\eta} - J_{\eta'}\|_{ker,\beta}\leq C_J\|\eta - \eta'\|_{tv,\beta}
		\end{equation}
		for some constant $C_J>0$ and that
		\begin{equation}\label{eq:uniform_fl}
			K_\eta V(x) \leq \wtilde{a}V(x) + \wtilde{b}
		\end{equation}
		for $\wtilde{a}\in ]0,1[,~\wtilde{b}>0$ and all $\eta\in \calP_{m_G,M_U}$ for suitably chosen constants $m_G,M_U$. Suppose $\delta$ is from \eqref{eq:delta_defn} and set
		\begin{description}
			\item[Case 1:] $\rho:=(1-\veps)\gamma$ if $J_\eta(x,\dd y)$ doesn't depend on $x$; or 
			\item[Case 2:] $\rho:=(1-\veps)\gamma + \veps\|J_{\eta^\star}V\|_V$ if $J_\eta(x,\dd y)$ does depend on $x$. 
		\end{description}
		If $\mu_0(V),\eta_0(U)<\infty$ then there exists a constant $C>0$ s.t.
		\[
			\|\mu_n - \pi \|_{tv,\beta}\leq \rho^n \|\mu_0 - \pi\|_{tv,\beta} + Cn\max(\rho,\delta)^n.
		\]In particular, if $\rho<1$, we have $\lim_{n\to \infty}\mu_n=\pi$ in $V_\beta$-total variation.
	\end{restatable}

	\subsection{Results About Weighted Total Variation }

	\begin{lemma}\label{lemma:conversion}
		Suppose that a Markov kernel $P$ satisfies a drift condition
		\[
			PV(x) \leq aV(x) + b
		\]for $V:\R^d\to [0,\infty[$, $V(x)\to \infty$ as $\|x\|\to \infty$, and the minorization condition with $\eps\in ]0,1[,~\nu\in \calP(\R^d)$, and $\calC:=\{x\st V(x)\leq R\}$ s.t.
		\[
			\inf_{x\in \calC} P(x,A)\geq \eps\nu(A).
		\]holds for some $R>2K/(1-a)$. Then there exists $\ovl{a}\in ]a,1[,~\ovl{b}>0,~\ovl{\nu}\in \calP(\R^d),~S\in \scrB(\R^d)$ s.t. 
		\[
			PV(x) \leq \ovl{a}V(x) + b\ind{x\in S},~~~\inf_{x\in S}P(x,A)\geq \ovl{\eps}~\ovl{\nu}(A).
		\]
	\end{lemma}
	\begin{proof}
		This argument merely an elaboration on the remark from the end of \cite{hairer2011yet}. Let $\ovl{a}\in ]a,1[$ such that $R>b/(\ovl{a}-a)$ and $S=\{x\st V(x)\leq R\}$. If $x\notin S$ then
		\begin{align*}
			PV(x) \leq aV(x) + b &= aV(x) + \frac{b}{V(x)}V(x)\leq aV(x) + \frac{b}{R}V(x)\\
			&\leq aV(x) + b\frac{\ovl{a}-a}{b}V(x)= aV(x) + (\ovl{a}-a)V(x) \\
			&= \ovl{a}V(x).
		\end{align*}
		If $x\in S$ then $PV(x)\leq aV(x) + b\leq \ovl{a}V(x) + b$. Hence the choices are clear from fixing $R$ as above. 
	\end{proof}

	\begin{lemma}\label{lem:v_tv}
		Let $P$ be a Markov kernel with invariant measure $\pi$ and suppose there are constants $\rho\in ]0,1[,~C>0$, and $r\in ]0,1]$ s.t.
		\[
			\|P^nf - \pi(f)\|_{V^r}\leq C\rho^n\|f\|_{V^r}
		\]for any $f:\R^d\to \R$ with $\|f\|_{V^r}<\infty$. Then if $\mu_0(V^r)<\infty$ and $\mu_n$ is the flow of $P$ then there is a constant $C'>0$ s.t.
		\[
			\|\mu_n - \pi\|_{tv,V^r}\leq C'\rho^n.
		\]
	\end{lemma}
	\begin{proof}
		Consider
		\begin{align*}
			\|\mu_n - \pi\|_{tv,V^r} &= \sup_{\|f\|_{V^r}\leq 1}|\mu_0P^n(f) - \pi(f)| \\
			&= \sup_{\|f\|_{V^r}\leq 1}|\mu_0(P^nf - \pi(f))|\\
			&\leq \sup_{\|f\|_{V^r}\leq 1}\mu_0(|P^nf(x) - \pi(f)|)\\
			&\leq \sup_{\|f\|_{V^r}\leq 1}\mu_0(V^r(x)\|P^nf - \pi(f)\|_{V^r})\\
			&\leq C\rho^n \mu_{0}(V^r)= C'\rho^n.
		\end{align*}
	\end{proof}

	\begin{lemma}\label{lem:weighted_bounds}
		Let $P,Q$ be Markov kernels and suppose that $P$ satisfies a drift condition
		\[
			PV(x) \leq aV(x) + b
		\]for $V:\R^d\to [1,\infty[$. Then
		\[
			\|\mu P - \nu P\|_{tv,V}\leq (a+b)\|\mu-\nu\|_{tv,V}
		\]and
		\[
			\|\mu P - \mu Q\|_{tv,V}\leq \mu(V)\|P-Q\|_{ker,V}.
		\]
	\end{lemma}
	\begin{proof}
		For the first part:
		\[
			\|\mu P - \nu P\|_{tv,V} = \sup_{\|f\|_{V}\leq 1}|\mu P(f) - \nu P(f)|\leq \sup_{\|f\|_{V},\|h\|_V\leq 1}\| Pf\|_{V}|\mu(h) - \nu(h)|
		\]and 
		\begin{align*}
			\|Pf\|_V &= \sup_x \frac{|Pf(x)|}{V(x)}
			= \sup_x \frac{|\int P(x,\dd y)f(y)|}{V(x)} 
			= \sup_x \frac{|\int P(x, \dd y)V(y)[f(y)/V(y)]}{V(x)}\\
			&\leq \|f\|_V \sup_x \frac{|PV(x)|}{V(x)}\leq \sup_x \frac{aV(x) + b}{V(x)}\leq a + b.
		\end{align*}
		For the second part, let $\|f\|_V\leq 1$
		\[
			|\mu P(f) - \mu Q(f)|\leq \mu(|Pf - Qf|)\leq \mu(V\frac{|Pf-Qf|}{V})\leq \mu(V)\|Pf-Qf\|_V\leq \mu(V)\|P-Q\|_{ker,V}.
		\]
	\end{proof}

	\subsection{Proof of Theorem~\ref{thm:gen_conv}}
	\genconv* 
	\begin{proof}
		Case 1: For $n=1$, consider
		\begin{align*}
			\|\mu_1 - \pi\|_{tv,\beta}&= \|\mu_0 K_{\eta_1} - \pi K_{\eta^\star}\|_{tv,\beta}\\
			&\leq (1-\veps)\|\mu_0 K - \pi K\|_{tv,\beta} + \veps\|\mu_0 J_{\eta_1} - \pi J_{\eta^\star}\|_{tv,\beta}\\
			&= (1-\veps)\|\mu_0 K - \pi K\|_{tv,\beta} + \veps\|J_{\eta_1} - J_{\eta^\star}\|_{tv,\beta}\\
			&\leq (1-\veps)\gamma \|\mu_0 - \pi\|_{tv,\beta} + \veps C_J\|\eta_1 - \eta^\star\|_{tv,\beta}\\
			&\leq \rho \|\mu_0 - \pi\|_{tv,\beta} + \veps C_JM(r^*)\delta \\
			&\leq \rho \|\mu_0 - \pi\|_{tv,\beta} + C\max(\rho,\delta)
		\end{align*}
		because from Assumption~\ref{assump:compat}-C1 we have \footnote{this is because $\|f\|_{U^{r^*}}\leq \|f\|_\beta\implies \|\mu - \nu\|_{tv,\beta}\leq \|\mu - \nu\|_{tv,U^{r^*}}$}
		\[
			\|\eta - \eta'\|_{tv,\beta}\leq \|\eta - \eta'\|_{tv,U^{r^*}}.
		\]Hence assume true for $n-1$. Then for $n$:
		\begin{align*}
			\|\mu_n - \pi\|_{tv,\beta} &\leq (1-\veps)\|\mu_{n-1}K - \pi K\|_{tv,\beta} + \veps\|\mu_{n-1} J_{\eta_n} - \pi J_{\eta^\star}\|_{tv,\beta}\\
			&= (1-\veps)\|\mu_{n-1}K - \pi K\|_{tv,\beta} + \veps\|J_{\eta_n} - J_{\eta^\star}\|_{tv,\beta}\\
			&\leq (1-\veps)\gamma\|\mu_{n-1} - \pi\|_{tv,\beta} + \veps C_J\|\eta_n - \eta^\star\|_{tv,\beta}\\
			&\leq \rho (\rho^{n-1}\|\mu_0 - \pi\|_{tv,\beta} + C(n-1)\max(\rho,\delta)^{n-1}) + \veps C_JM(r^*)\delta^n \\
			&\leq \rho^n \|\mu_0 - \pi\|_{tv,\beta} + C[(n-1)\rho \max(\rho,\delta)^{n-1} + \max(\rho,\delta)^n]\\
			&\leq \rho^n \|\mu_0 - \pi\|_{tv,\beta} + C[(n-1)\max(\rho,\delta)^n + \max(\rho,\delta)^n]\\
			&= \rho^n \|\mu_0 - \pi\|_{tv,\beta} + Cn\max(\rho,\delta)^n.
		\end{align*}
		hence Case 1 is done. 

		Case 2: We proceed the same way as Case 1 except we need to use
		\begin{align*}
			\|\mu_{n-1}J_{\eta_n} - \pi J_{\eta^\star}\|_{tv,\beta}&\leq \|\mu_{n-1}J_{\eta_n} - \mu_{n-1}J_{\eta^\star}\|_{tv,\beta} + \|\mu_{n-1}J_{\eta^\star} - \pi J_{\eta^\star}\|_{tv,\beta}\\
			&\leq \mu_{n-1}(V_\beta)\|J_{\eta_n} - J_{\eta^\star}\|_{tv,\beta} + \|J_{\eta^\star}V_\beta\|_{\beta}\|\mu_{n-1} - \pi\|_{tv,\beta}
		\end{align*}
		using Lemma~\ref{lem:weighted_bounds}. Note that since $K_\bdot$ satisfies a uniform drift criterion for $V$, using linearity we have
		\[
			\mu_{n}(V_\beta) \leq \ovl{a}^n\mu_0(V_\beta) + \frac{b}{1-\ovl{a}}\leq \mu_0(V_\beta) + \frac{b}{1-\ovl{a}}=:C_0.
		\]
		Hence for $n=1$:
		\begin{align*}
			\|\mu_1 - \pi\|_{tv,\beta}&= \|\mu_0 K_{\eta_1} - \pi K_{\eta^\star}\|_{tv,\beta}\\
			&\leq (1-\veps)\|\mu_0 K - \pi K\|_{tv,\beta} + \veps\|\mu_0 J_{\eta_1} - \pi J_{\eta^\star}\|_{tv,\beta}\\
			&\leq (1-\veps)\|\mu_0 K - \pi K\|_{tv,\beta} + \veps\mu_{0}(V_\beta)\|J_{\eta_1} - J_{\eta^\star}\|_{tv,\beta} + \veps\|J_{\eta^\star}V_\beta\|_\beta \|\mu_0 - \pi\|_{tv,\beta}\\
			&\leq [(1-\veps)\gamma + \veps \|J_{\eta^\star}V_\beta\|_{\beta}]\|\mu_0 - \pi\|_{tv,\beta} + \veps C_0C_J\|\eta_1 - \eta^\star\|_{tv,\beta}\\
			&\leq \rho \|\mu_0 - \pi\|_{tv,V} + C\max(\rho,\delta).
		\end{align*}
		Hence assume true for $n-1$. Then for $n$:
		\begin{align*}
			\|\mu_n - \pi\|_{tv,\beta} &\leq (1-\veps)\|\mu_{n-1}K - \pi K\|_{tv,\beta} + \veps\|\mu_{n-1} J_{\eta_n} - \pi J_{\eta^\star}\|_{tv,\beta}\\
			&\leq (1-\veps)\|\mu_{n-1}K - \pi K\|_{tv,\beta} + \veps\mu_{n-1}(V_\beta)\|J_{\eta_n} - J_{\eta^\star}\|_{tv,\beta} + \veps\|J_{\eta^\star}V_\beta\|_\beta \|\mu_{n-1} - \pi\|_{tv,\beta}\\
			&\leq (1-\veps)\gamma\|\mu_{n-1} - \pi\|_{tv,\beta} + \veps\mu_{n-1}(V_\beta)\|J_{\eta_n} - J_{\eta^\star}\|_{tv,\beta} + \veps\|J_{\eta^\star}V_\beta\|_\beta \|\mu_{n-1} - \pi\|_{tv,\beta}\\
			&\leq \rho (\rho^{n-1}\|\mu_0 - \pi\|_{tv,\beta} + C(n-1)\max(\rho,\delta)^{n-1}) + \veps C_0C_JM(r^*)\delta^n \\
			&\leq \rho^n \|\mu_0 - \pi\|_{tv,\beta} + C[(n-1)\rho \max(\rho,\delta)^{n-1} + \max(\rho,\delta)^n]\\
			&\leq \rho^n \|\mu_0 - \pi\|_{tv,\beta} + C[(n-1)\max(\rho,\delta)^n + \max(\rho,\delta)^n]\\
			&= \rho^n \|\mu_0 - \pi\|_{tv,\beta} + Cn\max(\rho,\delta)^n.
		\end{align*}
		hence Case 2 is done.
	\end{proof}

	\section{Large-Particle Convergence}\label{app:poc}
	We study the behaviour of the interacting particle system \eqref{eq:general_IPS} as $N\to \infty$. The behaviour of this system is probabilistically different from that of the mean field system \eqref{eq:gen_mf_system} because the particles in \eqref{eq:general_IPS} are \emph{coupled} whereas the particles in \eqref{eq:gen_mf_system} are \emph{independent}. The fact that we recover independence in the limit of a collection of interchangeable particles is a remarkable feature of interacting particle systems \citep{sznitman1991topics}.\nnewline

	\subsection{Result}
	For a collection of $N$ particles, under suitable assumptions on $K_\bdot$, we will show that any fixed-size $q$-block of particles of $\{X^1_n,\dots,X^q_n\}$ becomes independent as $N\to \infty$, and moreover this trend towards independence happens \emph{uniformly} in time. This phenomenon is called the uniform \emph{propagation of chaos} property\footnote{here, ``chaos'' is synonymous with ``statistical independence'', coming from the statistical physics intuition that a collection of independent particles are maximally disordered, or chaotic. This means that particles which start chaotic will approximately ``propagate their chaos'' through time despite interactions between the particles}.\nnewline

	Let us describe the dynamics of the distribution of the IPS \eqref{eq:general_IPS}. From \eqref{eq:general_IPS}, each $X^i_n$ evolves according to
	\[
		X^i_n \sim K_{m(Y_n)}(X^i_{n-1},\bdot)
	\]which indicates that, given $Y_n:=(Y^1_n,\dots,Y^N_n)$, $X^i_n$ is sampled independently. Letting ${X^{q,N}_n:=(X^1_n,\dots,X^q_n)}$ and $\mu^{q,N}_n:=\law(X^{q,N}_n)\in \calP((\R^d)^q)$, for measurable $f:(\R^d)^q\to \R$ one has
	\[
		\mu^{q,N}_n(f) = \E[f(X^{q,N}_n)] = \E[\E[f(X^{q,N}_n)|Y_n]] = \E[\mu^{q,N}_{n-1}K_{m(Y_n)}^{\otimes q}(f)]
	\]where the expectation is taken over the distribution of $Y_n=(Y^1_n,\dots,Y^N_n)$, which is $\eta_n^{\otimes N}$. We will use this decomposition to derive a uniform propagation of chaos result in the following theorem.

	\begin{restatable}{theorem}{pocgeneral}\label{thm:poc_general}
		Let $N\in \N,~q\in \{1,\dots,N\}$, and consider the interacting particle system $X_n:=(X^1_n, \dots, X^N_n),~Y_n:=(Y^1_n,\dots,Y^N_n)$ from \eqref{eq:general_IPS}.
		Let $\mu^{q,N}_n:=\law(X^1_n,\dots,X^q_n)$, and let $\mu_n$ be the distribution of the (independent) mean field system $\eqref{eq:gen_mf_system}$, with $X^i_0\sim \mu_0$. Suppose that $\forall x\in (\R^d)^q$, $\eta\in \calP(\R^d)$, and $f\in \calB_b((\R^d)^q)$ with $osc(f)\leq 1$\footnote{due to the characterization $\|\mu - \nu\|_{tv}=\sup\{|\mu(f) - \nu(f)|\st f\in \calB_b(\R^d),~\osc(f)\leq 1\}$ from \cite{del2004feynman} this regularity condition should be interpreted as a total variation Lipschitzness analogous to \eqref{eq:gen_reg}.}
		\[
			\left|\E[J^{\otimes q}_{m(Y)}f(x)] - J^{\otimes q}_{\eta}f(x)\right
			| \leq c\frac{q^2}{N}\calR(q^2/N),~\text{where}~Y=\{Y^1,\dots,Y^N\},~Y^i\sim \eta.
		\]Suppose finally that $\mu^{q,N}_0=\mu^{\otimes q}_0$ for any $1\leq q\leq N$.

		If \textbf{\emph{(Case 1)}}: $J_\eta(x,\bdot)$ doesn't depend on $x$, or if \textbf{\emph{(Case 2)}}: $J_\eta(x,\bdot)$ does depend on $x$ but additionally that $\eps(K)<1$, then there exists a fixed constant $C>0$ s.t.
		\[
			\sup_{n\geq 0}\|\mu^{q,N}_n - \mu_n^{\otimes q}\|_{tv} \leq C \frac{q^2}{N}\calR(q^2/N).
		\]
	\end{restatable}
	\begin{proof}
		\textbf{Case 1}: We first claim that 
		\[
			\|\mu^{q,N}_{n+1} - \mu^{\otimes q}_{n+1}\|_{tv}\leq c\veps\sum_{j=0}^{n-1}(1-\veps)^j\eps(K^{\otimes q})^j\cdot\frac{q^2}{N}\calR(q^2/N).
		\]Let $f\in \calB_b((\R^d)^q)$ s.t. $osc(f)\leq 1$. 

		Consider the following expression for general $n$:
		\begin{align*}
			\left|\mu^{q,N}_{n+1}(f) - \mu^{\otimes q}_{n+1}(f)\right|&=\left|\E[\mu^{q,N}_n K^{\otimes q}_{m(Y_n)}(f)] - \mu^{\otimes q}_n K^{\otimes q}_{\eta_n}(f)\right|\\
			&= \left|\E[\mu^{q,N}_n K^{\otimes q}_{m(Y_n)}(f)] - \mu^{q,N}_n K^{\otimes q}_{\eta_n}(f) + \mu^{q,N}_n K^{\otimes q}_{\eta_n} (f)- \mu^{\otimes q}_n K^{\otimes q}_{\eta_n}(f)\right|\\
			&\leq \left|\E[\mu^{q,N}_n K^{\otimes q}_{m(Y_n)}(f)] - \mu^{q,N}_n K^{\otimes q}_{\eta_n}(f)\right| + \left|\mu^{q,N}_n K^{\otimes q}_{\eta_n} (f)- \mu^{\otimes q}_n K^{\otimes q}_{\eta_n}(f)\right|\\
			&\leq \sup_x \veps \left|\E[J_{m(Y_n)}^{\otimes q}f(x)] - J_{\eta_n}^{\otimes q}f(x)\right| + (1-\veps)\left|\mu^{q,N}_n K^{\otimes q}(f)- \mu^{\otimes q}_n K^{\otimes q} (f)\right|\\
			&\leq c\veps\frac{q^2}{N}\calR(q^2/N) + (1-\veps)\eps(K^{\otimes q})\left|\mu^{q,N}_n(f) - \mu^{\otimes q}_n(f)\right|.
		\end{align*}
		where we have used the assumption on $J$. 

		Thus if $n=1$ in the above expression, the base case holds since $\mu^{q,N}_0 = \mu^{\otimes q}_0$. Now, if the claim holds for $n$, then for $n+1$ we have
		\begin{align*}
			|\mu^{q,N}_{n+1}(f) - \mu^{\otimes q}_{n+1}(f)|&\leq c\veps\frac{q^2}{N}\calR(q^2/N) + (1-\veps)\eps(K^{\otimes q})c\veps\sum_{j=0}^{n-1}(1-\veps)^j\eps(K^{\otimes q})^j\cdot\frac{q^2}{N}\calR(q^2/N)\\
			&= c\veps\frac{q^2}{N}\calR(q^2/N)\left[1 + (1-\veps)\eps(K^{\otimes q})\sum_{j=0}^{n-1}(1-\veps)^j\eps(K^{\otimes q})^j\right]\\
			&= c\veps\frac{q^2}{N}\calR(q^2/N)\sum_{j=0}^n(1-\veps)^j\eps(K^{\otimes q})^j.
		\end{align*}
		Thus the claim holds for $n+1$. Since $(1-\veps)\eps(K^{\otimes q})<1$, we have the result.

		\textbf{Case 2}: We first claim that
		\[
			\|\mu^{q,N}_{n+1} - \mu^{\otimes q}_{n+1}\|_{tv}\leq c\veps\sum_{j=0}^{n-1}[\veps + (1-\veps)\eps(K^{\otimes q})]^j\cdot\frac{q^2}{N}\calR(q^2/N).
		\]
		Let $f\in \calB_b((\R^d)^q)$ with $osc(f)\leq 1$. We proceed by induction. Note that:
		\begin{align*}
			\left|\mu^{q,N}_{n+1}(f) - \mu^{\otimes q}_{n+1}(f)\right|&=\left|\E[\mu^{q,N}_n K^{\otimes q}_{m(Y_n)}(f)] - \mu^{\otimes q}_n K^{\otimes q}_{\eta_n}(f)\right|\\
			& = \veps\left|\E[\mu^{q,N}_n J^{\otimes q}_{m(Y_n)}(f) ]- \mu^{\otimes q}_n J^{\otimes q}_{\eta_n}(f)\right| + (1-\veps)\left|\mu^{q,N}_nK(f) - \mu^{\otimes q}_nK(f)\right|
		\end{align*}
		for the first term, we will use the decomposition
		\[
			\left|\E[\mu^{q,N}_n J_{m(Y_n)}(f)]- \mu^{\otimes q}_n J_{\eta_n}(f)\right| \leq \left|\E[\mu^{q,N}_n J_{m(Y_n)}(f) ]- \mu^{q,N}_n J_{\eta_n}(f)\right| + \left|\mu^{q,N}_n J_{\eta_n}(f)- \mu^{\otimes q}_n J_{\eta_n}(f)\right|
		\]Then, using the assumption on $J$, we have
		\[
			\left|\E[\mu^{q,N}_n J_{m(Y_n)}(f) ]- \mu^{q,N}_n J_{\eta_n}(f)\right|\leq c\frac{q^2}{N}\calR(q^2/N))
		\]and also we know we can use
		\begin{align*}
			\left|\mu^{q,N}_nJ_{\eta_n}(f) - \mu^{\otimes q}_nJ_{\eta_n}(f)\right|&\leq \osc(J_{\eta_n}f)\left|\mu^{q,N}_n(h) - \mu^{\otimes q}_n(h)\right|\\
			&\leq \eps(J_{\eta_n})\osc(f)\left|\mu^{q,N}_n(h) - \mu^{\otimes q}_n(h)\right|\\
			&\leq \left|\mu^{q,N}_n(h) - \mu^{\otimes q}_n(h)\right|
		\end{align*}
		since $\eps(J_{\eta_n})\leq 1$, with $\osc(h)\leq 1$. Hence putting these together, we see
		\begin{align*}
			\left|\mu^{q,N}_{n+1}(f) - \mu^{\otimes q}_{n+1}(f)\right| &\leq \veps c\frac{q^2}{N}\calR(q^2/2N) + \veps\left|\mu^{q,N}_n(h) - \mu^{\otimes q}_n(h)\right| + (1-\veps)\eps(K^{\otimes q})\left|\mu^{q,N}_n(f) - \mu^{\otimes q}_n(f)\right|\\
			&\leq \veps c\frac{q^2}{N}\calR(q^2/2N) + (\veps + (1-\veps)\eps(K^{\otimes q}))\|\mu^{q,N}_n - \mu^{\otimes q}_n\|_{tv}.
		\end{align*}

		Now if $n=0$ in the above expression, the base case holds since $\mu^{q,N}_0=\mu_0^{\otimes q}$. If it is true for $n$, then for $n+1$ we have
		\begin{align*}
			&\left|\mu^{q,N}_{n+1}(f) - \mu^{\otimes q}_{n+1}(f)\right|=\left|\E[\mu^{q,N}_nK_{m(Y_n)}(f)] - \mu^{\otimes q}_nK_{\eta_n}(f)\right|\\
			&\leq c\veps\frac{q^2}{N}\calR(q^2/N) + (\veps + (1-\veps)\eps(K^{\otimes q}))\|\mu^{q,N}_n(f) - \mu^{\otimes q}_n(f)\|_{tv}\\
			&\leq c\veps\frac{q^2}{N}\calR(q^2/N) + (\veps + (1-\veps)\eps(K^{\otimes q}))\left[c\veps\sum_{j=0}^{n-1}[\veps + (1-\veps)\eps(K^{\otimes q})]^j\cdot\frac{q^2}{N}\calR(2q^2/N)\right]\\
			&= c\veps\frac{q^2}{N}\calR(q^2/N) \left[1 + \sum_{j=1}^{n}[\veps + (1-\veps)\eps(K^{\otimes q})]^j\right]\\ 
			&=c\veps\frac{q^2}{N}\calR(q^2/N)\sum_{j=0}^{n}[\veps + (1-\veps)\eps(K^{\otimes q})]^j. 
		\end{align*}
		Hence if $\eps(K^{\otimes q})<1$, i.e. $\eps(K)<1$, the result holds.
	\end{proof}

	\subsection{Proof of Main Results (Theorem~\ref{thm:main_result} \& Corollary \ref{coro:mc})}\label{app:main_proofs}
	\mainthm* 
	\begin{proof}
		This follows straightforwardly from the above discussion, the only technicality is converting the results from Theorem~\ref{thm:gen_conv} to the un-weighted total variation. But note that since $\{\|f\|_\infty\leq 1\}\subset \{\|f\|_\beta\leq 1\}$ we have
		\[
			\|\mu_n - \mu_\infty\|_{tv} = \sup_{\|f\|_\infty\leq 1}|\mu_n(f) - \mu_\infty(f)|\leq \sup_{\|f\|_\beta\leq 1}|\mu_n(f) - \mu_\infty(f)| = \|\mu_n - \mu_\infty\|_{tv,\beta}
		\]so we're done.
	\end{proof}

	\maincoro* 
	\begin{proof}
		Let $f\in \calB_b(\R^d)$ and consider
		\begin{align*}
			\E[(m(X_n)(f) - \mu_n(f))^2]&= \E\left[\left(\frac{1}{N}\sum_{i=1}^N f(X^i_n) - \mu_n(f)\right)^2\right]\\
			&= \frac{1}{N^2}\sum_{i=1}^n \E[f(X^i_n)f(X^j_n)] - \frac{2}{N}\sum_{i=1}^N \E[f(X^i_n)]\mu_n(f) + \mu_n(f)^2\\
			&= \frac{1}{N}\E[f(X^1_n)^2] + \frac{N-1}{N}\E[f(X^1_n)f(X^2_n)] - 2\E[f(X^1_n)]\mu_n(f) + \mu_n(f)^2
		\end{align*}
		using interchangeability of the $X^i_n$s. Since the strong propagation of chaos in Theorem~\ref{thm:poc_general} clearly implies weak propagation of chaos, i.e. for any bounded $f$ $|\mu^{q,N}_n(f) - \mu^{\otimes q}_n(f)|\to 0$, we see that the expression above $\to 0$ and we have $L^2$-convergence which implies weak convergence.
	\end{proof}

	\section{Analysis of the Kernels $K^{BG}_\eta$ and $K^{AR}_\eta$}\label{app:specific}
	\subsection{Analysis of $K^{BG}_\eta$}\label{app:specific_bg}
	\subsubsection{Statement of Results}
	A key difficulty that arises when working with $\Psi_G(\eta)=\frac{G\dd \eta}{\eta(G)}$ is obtaining a uniform lower bound the denominator $\eta(G)$. This is important for deriving uniform regularity results and for generally establishing conditions under which the measures $\{\Psi_G(\eta_n)\}_{n=0}^\infty$ are well-defined. An effective approach used in \cite{del2004feynman} and related works is to assume the uniform lower bound $G(x)>\eps>0~\forall x\in \R^d$ for some $\eps>0$. While this assumption is self-contained in that it works for any choice of $\eta$, it eliminates ubiquitous families of measures such as Gaussians, e.g. $G(x)=\exp(-\|x\|^2/2)$ and $\dd\eta=\dd x$. \nnewline

	We will see below how to relax $G(x)>\veps > 0$ using the structure of our problem. In particular, we can use the Lyapunov function $U$ for $Q$ to control the probability that the state $Y_n\sim \eta_n$ will venture ``far away'' from the center of the state space. This leads to the intuition that, if $G$ is sufficiently ``compatible'' with $U$, i.e. the function $G$ stays away from zero in the ``center'' of the state space as determined by $U$, then the expectation $\eta_n(G)=\E_{Y_n\sim \eta_n}[G(Y_n)]$ can be lower-bounded using the same Lyapunov function. This insight is encoded in the compatibility criterion Assumption~\ref{assump:compat}-C2, and we can use it to prove the following \emph{a priori} lower bound on $\eta_n(G)$.

	\begin{restatable}{lemma}{uniflb}\label{lem:unif_lb}
		Suppose that Assumption~\ref{assump:FL}-Q1 and Assumption~\ref{assump:compat}-C2 hold, and let $\eta_n=\eta_{n-1}Q$ with $\eta_0(U)<\infty$. Then we have the lower bound
		\[
			\eta_n(G) \geq \theta(R^*)\left(1 - \frac{\gamma^n\eta_0(U) + \frac{c}{1-\gamma}}{R^*}\right)
		\]where $R^*$ is fixed and doesn't depend on $n$.
	\end{restatable}
	This lemma essentially says that $G$ should be bounded away from zero on the level sets of $U$. This is not a strong condition --- if $G$ is bounded away from zero on compact sets and $U$ is continuous, the lemma applies. As an example, if $G=\exp(-\|x\|^2/2)$ and $U(x)=c\|x\|^2 + 1$ then this result applies. The constant $R^*$ arises since $R\mapsto \theta(R)$ is nonincreasing and $R\mapsto 1-\frac{1}{R}$ is increasing so we can optimize this bound as a function of $R$ to get $R^*$ (which may not be unique, but the value the bound attains will be). Together with the next lemma, we can obtain the desired convergence from Theorem~\ref{thm:gen_conv} for the BG interaction.

	\begin{restatable}{lemma}{psireg}\label{lem:psi_reg}
		Let $\eta, \eta'\in \calP(\R^d)$, and suppose that Assumption~\ref{assump:G} holds, i.e. $\|G\|_\infty,\|G\|_\beta<\infty$. Then
		\[
			\|\Psi_G(\eta) - \Psi_G(\eta')\|_{tv,\beta}\leq \left(\frac{\|G\|_{\beta} + \|G\|_\infty}{\eta(G)\vee \eta'(G)} + \eta(V)\wedge \eta'(V)\beta\frac{\|G\|_\beta \|G\|_\infty}{\eta(G)\eta'(G)}\right)\|\eta-\eta'\|_{tv,\beta}.
		\]
	\end{restatable}

	Clearly, we will use our knowledge that $\eta_n\in \calP_{m_G,M_U}(\R^d)~\forall n$ to make the Lipschitz constant in Lemma~\ref{lem:psi_reg} uniform over $\calP_{m_G,M_U}$. 

	\begin{restatable}{proposition}{bgunfiormfl}\label{prop:bg_uniform_fl}
		Suppose that Assumption~\ref{assump:FL},~\ref{assump:compat}, and Assumption~\ref{assump:G}-G1 hold. Then the drift criterion \eqref{eq:uniform_fl} holds uniformly over all $\eta\in \calP_{m,M}(\R^d)$.
	\end{restatable}

	Now we can apply Theorem~\ref{thm:gen_conv} to obtain convergence.

	\begin{corollary}\label{coro:bg_conv}
		Suppose that Assumptions~\ref{assump:FL},~\ref{assump:compat},~\ref{assump:G} hold. If $\eta_0(G)>0$, $\eta_0(U),\mu_0(V)<\infty$, then Theorem~\ref{thm:gen_conv} holds for $K^{BG}$, i.e. the flow $\mu_n$ converges to $\pi$ in $V_\beta$-total variation as long as $\rho=(1-\veps)\gamma<1$.
	\end{corollary}
	 \begin{proof}
	 	Lemma~\ref{lem:unif_lb} and Lemma~\ref{lem:psi_reg} imply the regularity condition \eqref{eq:gen_reg} for constants determined by those lemmas and by noting that, due to Asssumption~\ref{assump:compat}-C1, $\eta(V_\beta)\leq \eta(U^{r^*})\leq \eta(U)^{r^*}<\infty$ by Jensen's inequality. Additionally, Proposition~\ref{prop:bg_uniform_fl} implies the condition \eqref{eq:uniform_fl}. Hence, since we are in ``Case 1'' of Theorem~\ref{thm:gen_conv}, the result follows.
	 \end{proof}

	 \subsubsection{Proofs}

	 \uniflb* 
	\begin{proof}
		\begin{align*}
			QG(x)&=\E_{X\sim Q(x,\bdot)}[G(X)] \\
			&= \E_{X\sim Q(x,\bdot)}[G(X)\ind{U(X)\leq R}] + \E_{X\sim Q(x,\bdot)}[G(X)\ind{U(X)>R}]\\
			&\geq \theta(R)\cdot \P_x(U(X)\leq R) + \E_{X\sim Q(x,\bdot)}[G(X)\ind{U(X)>R}]\\
			&\geq \theta(R)\cdot \P_x(U(X)\leq R).
		\end{align*}
		But by Markov's inequality
		\[
			\P_x(U(X)>R)\leq \frac{\E_x[U(X)]}{R} \leq \frac{\xi U(x) + c}{R}
		\]so 
		\[
			\P_x(U(X)\leq R) = 1 - \P_x(U(X)>R) \geq 1 - \frac{\xi U(x) + c}{R}
		\]and hence
		\[
			QG(x) \geq \theta(R) \P_x(U(X)\leq R) \geq \theta(R)\left(1 - \frac{\xi U(x) + c}{R}\right).
		\]Now 
		\begin{align*}
			Q^2G(x) &= Q(QG(x)) \geq Q\left(\theta(R)\left(1 - \frac{\xi U(x) + c}{R}\right)\right)\\
			&= \theta(R)\left(1 - \frac{\xi QU(x) + c}{R}\right)\geq \theta(R)\left(1 - \frac{\xi^2U(x) + c(1 + \xi)}{R}\right)
		\end{align*}
		and iterating this procedure gives 
		\[
			Q^nG(x)\geq \theta(R)\left(1 - \frac{\xi^nU(x) + \frac{c}{1-\xi}}{R}\right)
		\]where we have used the sum of the geometric series to obtain $c/(1-\xi)$. Now, $\theta(R)$ is nonincreasing w.r.t. $R$ and $1 - 1/R$ is increasing w.r.t. $R$, so optimizing to get $R^*$ and integrating we obtain
		\[
			\eta_n(G) = \eta_0(Q^nG) \geq \theta(R^*)\left(1 - \frac{\xi^n\eta_0(U) + \frac{c}{1-\xi}}{R^*}\right).
		\]
	\end{proof}

	\psireg* 
	\begin{proof}
		Let $\|f\|_\beta\leq 1$. then
		\begin{align*}
			&|\Psi_G(\eta)(f) - \Psi_G(\eta')(f)|=\left|\int \frac{G(x)f(x)}{\eta(G)}\eta(\dd x) - \int \frac{G(x)f(x)}{\eta'(G)}\eta'(\dd x)\right|\\
			&\leq \left|\int \frac{G(x)f(x)}{\eta(G)}\eta(\dd x) - \int \frac{G(x)f(x)}{\eta'(G)}\eta(\dd x)\right| + \left|\int \frac{G(x)f(x)}{\eta'(G)}\eta(\dd x) - \int \frac{G(x)f(x)}{\eta'(G)}\eta'(\dd x)\right|
		\end{align*}
		for the first term
		\begin{align*}
			&\left|\int \left(\frac{G(x)f(x)}{\eta(G)} - \frac{G(x)f(x)}{\eta'(G)}\right)\eta(\dd x)\right|\\
			&\leq \frac{|\eta(G) - \eta'(G)|}{\eta(G)\eta'(G)}\int G(x)|f(x)|\eta(\dd x)\\
			&\leq \frac{\|G\|_{\beta}\|\eta-\eta'\|_{tv,\beta}}{\eta(G)\eta'(G)}\int G(x)|f(x)|\eta(\dd x)\\
			&\leq \frac{\|G\|_{\beta}\|\eta-\eta'\|_{tv,\beta}}{\eta(G)\eta'(G)}\eta(G(1 + \beta V))\|f\|_\beta\\
			&= \left(\frac{\|G\|_{\beta}}{\eta'(G)} + \beta \eta(V)\frac{\|G\|_\beta \|G\|_\infty}{\eta(G)\eta'(G)}\right)\|\eta-\eta'\|_{tv,\beta}
		\end{align*}
		and for the second
		\begin{align*}
			&\left|\int \frac{G(x)f(x)}{\eta'(G)}\eta(\dd x) - \int \frac{G(x)f(x)}{\eta'(G)}\eta'(\dd x)\right|\\
			&=\frac{1}{\eta'(G)}\left|\int G(x)f(x)\eta(\dd x) - \int G(x)f(x)\eta'(\dd x)\right|\\
			&\leq \frac{\|fG\|_\beta}{\eta'(G)}\|\eta-\eta'\|_{tv,\beta}\\
			&\leq \frac{\|G\|_\infty\|f\|_\beta}{\eta'(G)}\|\eta-\eta'\|_{tv,\beta} = \frac{\|G\|_\infty}{\eta'(G)}\|\eta-\eta'\|_{tv,\beta}
		\end{align*}
		since\[
			\|fG\|_\beta = \sup_{x}\frac{|f(x)G(x)|}{1 + \beta V(x)}\leq \|G\|_\infty \|f\|_\beta
		\]so putting these together
		\[
			\|\Psi_G(\eta) - \Psi_G(\eta')\|_{tv,\beta}\leq \left(\frac{\|G\|_{\beta} + \|G\|_\infty}{\eta'(G)} + \eta(V)\beta\frac{\|G\|_\beta \|G\|_\infty}{\eta(G)\eta'(G)} \right)\|\eta-\eta'\|_{tv,\beta}
		\]Using symmetry completes the proof.
	\end{proof}

	\bgunfiormfl* 
	\begin{proof}
		Let $\eta\in \calP_{m,M}(\R^d)$ and consider
		\begin{align*}
			K_{\eta}V(x) & = (1-\veps)KV(x) + \veps \Psi_G(\eta)(V)\\
			&\leq (1-\veps)aV(x) + (1-\veps)b + \veps\frac{\eta(V)}{\eta(G)}\\
			&\leq (1-\veps)a V(x) + (1-\veps)b + \veps\frac{M}{m}.
		\end{align*}
	\end{proof}

	\begin{theorem}[\cite{del2004feynman} Thm 8.7.1 pp.283]\label{lem:poc_psi}
		Suppose that $G$ has bounded oscillations, let $N\geq q\geq 1$, and $Y=(Y^1,\dots,Y^N),~Y^i\sim \eta$. Then for any $f\in \calB_b((\R^d)^q)$ with $\osc(f)\leq 1$, we have
		\[
			|\E[\Psi_{G^{\otimes q}}(m(Y)^{\otimes q})(f) - \Psi_{G^{\otimes q}}(\eta^{\otimes q})(f)]|\leq c\frac{q^2}{N}\calR_{G,\eta}(2q^2/N)
		\]where
		\[
			\calR_{G,\eta}(u) := 1 + \osc_\eta(G)^2(1 + \osc_\eta(G)\sqrt{u})\exp(\osc_\eta(G)^2 u),~~~\osc_\eta(G):=\osc(G/\eta(G)).
		\] In particular, by picking $G\equiv 1$, we obtain
		\[
			|\E[m(Y)^{\otimes q}(f) - \eta^{\otimes q}(f)]|\leq c\frac{q^2}{N}.
		\]
	\end{theorem}

	\begin{corollary}\label{coro:bg_unif_poc}
		Suppose that Assumption~\ref{assump:G}~G1 holds (i.e. $G$ is bounded) the compatibility criterion Assumption~\ref{assump:compat}-C2 with $\eta_n(G)\geq m$.
		Then $K^{BG}_\bdot$ satisfies the uniform propagation of chaos in Case 1 with 
		\[
			\calR(u) = \calR_G(u):=1 + \frac{\osc(G)^2}{m^2}\left(1 + \frac{\osc(G)}{m}\sqrt{u}\right)\exp\left(\frac{\osc(G)^2}{m^2} u\right)
		\]
	\end{corollary}
	\begin{proof}
		This follows from Theorem~\ref{thm:poc_general} and Theorem~\ref{lem:poc_psi} above, noting 1) that $\|G\|_\infty$ implies $\osc(G)<\infty$, and 2) that we can obtain a bound on $\calR_{G,\eta_n}$ independent of $\eta_n$ by applying Lemma~\ref{lem:unif_lb} to obtain $\eta_n(G)\geq m$ so
		\[
			osc_{\eta_n}(G) = \osc(G/\eta_n(G))\leq \osc(G)/m.
		\]
	\end{proof}

	\subsection{Analysis of $K^{AR}_\eta$}

	\subsubsection{Statement of Results}
	First, the result showing that $\pi$ is $K^{AR}_{\eta^\star}$-invariant.
	\begin{proposition}\label{prop:ar_invar}
		$K_{\eta^\star}$ is $\pi$-invariant.
	\end{proposition}
	\begin{proof}
		We have
		\begin{align*}
			\pi(J_{\eta^\star}f) &= \iint [f(y) - f(x)] \alpha(x,y) \eta^\star(\dd y)\pi(\dd x) + \pi(f)\\
			&= \iint [f(y) - f(x)] 1\wedge \frac{\pi(y)\eta^\star(x)}{\eta^\star(y)\pi(x)}\eta^\star(\dd y)\pi(\dd x) + \pi(f)\\
			&= \iint [f(y) - f(x)]\pi(x)\eta^\star(y)\wedge \pi(y)\eta^\star(x)\dd y \dd x + \pi(f)\\
			&= \pi(f)
		\end{align*}
		since the integrals with $f(y)$ and $f(x)$ in the first term are equal.
	\end{proof}

	We will establish equvalent results from Section~\ref{app:specific_bg}but for  $K^{AR}_\bdot$. This will not require assumption~\ref{assump:compat}-C2 since the interaction is well-defined for $\eta\in \calP(\R^d)$ as $\alpha(x,y)$ is in fact \emph{bounded}. $K^{AR}$ is the main subject of study for \cite{delmoralnonlin}, and in fact the regularity and uniform drift conditions were established there. Note that the statement $\eta\in \calP_{0,\infty}$ is vaccuous since $G(x)>0\implies \eta(G)>0$, and the second says that $\eta(V)<\infty$. 

	\begin{lemma}[\cite{delmoralnonlin}]
		Let $\eta,\eta'\in \calP_{0,\infty}(\R^d)$. Then 
		\[
			\|J^{AR}_\eta - J^{AR}_{\eta'}\|_{ker,\beta}\leq 2\|\eta - \eta'\|_{tv,\beta}.
		\]
	\end{lemma}
	\begin{restatable}{lemma}{arunfiormfl}\label{prop:ar_uniform_fl}
		Suppose that Assumption~\ref{assump:FL},~\ref{assump:compat}-C2,~ hold. Then the uniform drift criterion \eqref{eq:uniform_fl} holds for $\eta\in \calP_{0,M}(\R^d)$.
	\end{restatable}

	\begin{restatable}{proposition}{arpoc}
		Let $J = J^{AR}$, and $Y=\{Y^1,\dots,Y^N\}$ where $Y^i\iid \eta$. Then for any $f\in \calB_b((\R^d)^q$ with $\osc(f)\leq 1$ and $x\in (\R^d)^q$, we have
		\[
			|\E[J^{\otimes q}_{m(Y)}f(x)] - J_\eta^{\otimes q}f(x)|\leq 2c\frac{q^2}{N}.
		\]
	\end{restatable}

	\begin{corollary}\label{coro:ar_conv}
		Suppose that Assumption~\ref{assump:FL},~\ref{assump:compat}-C2 hold. If $\eta_0(U),\mu_0(V)<\infty$, then Theorem~\ref{thm:gen_conv} holds for $K^{BG}$, i.e. the flow $\mu_n$ converges to $\pi$ in $V_\beta$-total variation as long as $\rho=(1-\veps)\gamma + \veps\|J_{\eta_0}V_\beta\|<1$.
	\end{corollary}

	\subsubsection{Proofs}
	\arunfiormfl* 
	\begin{proof}
		Let $\eta\in \calP_{0,M}(\R^d)$ and consider
		\begin{align*}
			K^{AR}_\eta V(x) & = (1-\veps)KV(x) + \veps [\eta(\alpha(x,\bdot)V) + (1-A_\eta(x))V(x)]\\
			&\leq (1-\veps)aV(x) + (1-\veps)b + \veps \eta(V) + \veps V(x)\\
			&\leq [(1-\veps)a + \veps]V(x) + (1-\veps)b + \veps M.
		\end{align*}
	\end{proof}

	\arpoc* 
	\begin{proof}
		Starting with the first term, for each fixed $x$
		\begin{align*}
			|\E[J^{\otimes q}_{m(Y_n)}f(x)] - J^{\otimes q}_{\eta_n}f(x)|& = \left|\E\left[\int [f(y) - f(x)]\alpha^{\otimes q}(x,y)m(Y_n)^{\otimes q}(\dd y)\right] - \int [f(y) - f(x)]\alpha^{\otimes q}(x,y)\eta^{\otimes q}_n(\dd y)\right|\\
			&= |\E[m(Y_n)^{\otimes q}(\vphi_f(x,\bdot))] - \eta^{\otimes q}_n(\vphi_f(x,\bdot))|
		\end{align*}
		where 
		\[
			\vphi_f(x, y) := \alpha^{\otimes q}(x,y)[f(y) - f(x)].
		\]Now
		\[
			\sup_y |\vphi_f(x, y)| = \sup_y|[f(y) - f(x)]\alpha^{\otimes q}(x,y)|\leq \sup_y |f(y)-f(x)|\|\alpha^{\otimes q}(x,\bdot)\|_{\infty}\leq \osc(f) \leq 1
		\]so automatically $\osc(\vphi_f(x,\bdot))\leq 2$. Hence using Lemma~\ref{lem:poc_psi}
		\[
			|\E[m(Y_n)^{\otimes q}(\vphi_f(x,\bdot))] - \eta^{\otimes q}_n(\vphi_f(x,\bdot))|\leq 2c \frac{q^2}{N}.
		\]
	\end{proof}

	\begin{corollary}\label{coro:ar_unif_poc}
		If $\eps(K)<1$, then $K^{AR}_\bdot$ satisfies the uniform propagation of chaos in Case 2 of Theorem~\ref{thm:poc_general} with $\calR\equiv 1$.
	\end{corollary}

\appendix

\end{document}